\documentclass[pra,floatfix,amsmath,superscriptaddress,twocolumn,showkeys]{revtex4}

\usepackage{amssymb}
\usepackage{exscale}
\usepackage{graphicx}
\usepackage{graphics}
\usepackage{amsmath}
\usepackage{amsthm}
\usepackage{color}
\usepackage{enumerate}
\usepackage[usenames,dvipsnames]{xcolor}
\usepackage{subfigure}
\usepackage{float}
\usepackage[mathscr]{eucal}
\usepackage{bm}
\usepackage{mathtools}
\usepackage{hyperref}

\def\bea{\begin{eqnarray}}
\def\eea{\end{eqnarray}}
\def\bi{\begin{itemize}}
	\def\ei{\end{itemize}}
\def\bc{\begin{center}}
	\def\ec{\end{center}}
\def\ba{\begin{aligned}}
	\def\ea{\end{aligned}}
\def\be{\begin{equation}}
\def\ee{\end{equation}}
\def\bestar{\begin{equation*}}
\def\eestar{\end{equation*}}
\def\bt{\begin{tabular}}
	\def\et{\end{tabular}}

\def\C{\hbox{$\mit I$\kern-.7em$\mit C$}} 
\def\R{\hbox{$\mit I$\kern-.6em$\mit R$}}

\def\etabar{\bar{\eta}}
\def\edge{\bm{e}}
\def\htilde{\tilde{h}}
\def\qstar{q_{*}}
\def\gammadis{\gamma_{\mathrm{dis}}}

\def\normw#1{{\left\Vert #1 \right\Vert}_\mathrm{W}}   

\newtheorem{theorem}{Theorem}

\DeclareMathOperator{\E}{E}
\DeclareMathOperator*{\argmax}{arg\,max}

\begin{document}

\author{W. L. Boyajian}
\author{J. Clausen}
\author{L. M. Trenkwalder}
\affiliation{Institute for Theoretical Physics, University of Innsbruck, 6020 Innsbruck, Austria}
\author{V. Dunjko}
\affiliation{Max Planck Institute of Quantum Optics, 85748 Garching, Germany}
\affiliation{LIACS, Leiden University, Niels Bohrweg 1, 2333 CA Leiden, The  Netherlands}
\author{H. J. Briegel}
\affiliation{Institute for Theoretical Physics, University of Innsbruck, 6020 Innsbruck, Austria}
\affiliation{Department of Philosophy, University of Konstanz,  78457 Konstanz, Germany}

\title{On the convergence of projective-simulation-based reinforcement learning in Markov decision processes}
\date{\today}

\begin{abstract}
\textbf{Abstract.}
In recent years, the interest in leveraging quantum effects for enhancing machine learning tasks has significantly increased. 
Many algorithms speeding up supervised and unsupervised learning were established. 
The first framework in which ways to exploit quantum resources specifically for the broader context of reinforcement learning were found is projective simulation. 
Projective simulation presents an agent-based reinforcement learning approach designed in a manner which may support quantum walk-based speed-ups. 
Although classical variants of projective simulation have been benchmarked against common reinforcement learning algorithms, very few formal theoretical analyses have been provided for its performance in standard learning scenarios. 
In this paper, we provide a detailed formal discussion of the properties of this model. 
Specifically, we prove that one version of the projective simulation model, understood as a reinforcement learning approach, converges to optimal behavior in a large class of Markov decision processes. 
This proof shows that a physically-inspired approach to reinforcement learning can guarantee to converge. 
\end{abstract}

\maketitle

%

\section{\label{sect:intro} Introduction}

In the past decade, quantum information science established itself as a fruitful research field that leverages quantum effects to enhance communication and information processing tasks {bookNielsen, Bennett1995}.
The results and insights gained inspired further investigations which more recently contributed to the emergence of the field quantum machine learning \cite{schuld20114QuestQNN, biamonte2016qml, dunjko2018aireview}.
The aim of this new field is twofold. 
On the one hand, machine learning methods are developed to further our understanding and control of physical systems and, on the other hand, quantum information processing is employed to enhance certain aspects of machine learning. 
A learning framework that features in both aspects of quantum machine learning is projective simulation (PS). 
In particular, PS can be seen as a platform for the design of autonomous (quantum) learning agents \cite{Bri12}.

The development of projective simulation is not motivated by the aim of designing ever-faster computer algorithms. 
Projective simulation is a tool for understanding various aspects of learning, where agents are viewed from the perspective of realizable entities such as robots or biological systems interacting with an unknown environment. 
In this embodied approach, the agent's perception is influenced by its sensors, its actions are limited by its physical capabilities and its memory is altered by its interaction with an environment. 
The deliberation process of the agent can be described by a random walk process on the memory structure and it is their quantum counterpart, quantum random walks, that offers a direct route to the quantization of the deliberation and learning process.
Thereby, PS not only allows us to study learning in the quantum domain it also offers speed-ups in a variety of learning settings \cite{Pap14, riarunothai2017speeding}.

Projective simulation can be used to solve reinforcement learning (RL) problems as well. Taken as a classical RL approach, the PS has proven to be a successful tool for learning how to design quantum experiments \cite{Mel18}.
In  \cite{Mel18} PS was used to design experiments that generate high-dimensional multipartite entangled photonic states. 
The ability of PS to learn and adapt to an unknown environment was further used for optimizing and adapting quantum error correction codes \cite{HPN19}.
In a quite different context, PS is used to model complex skill acquisition in robotics \cite{Han20, Han16}. 

Although PS has been shown suitable for a number of applications, it is a fair question of just how well it does, compared to other models, or compared to theoretical optima.
However, the empirical evaluation of a model through simulations, and analytically proving the properties of the same model are fundamentally distinct matters.
For example, in many applications, empirical convergence can be reached even if the conditions for theoretical convergence are not met.
In any real-world application, such as learning to play the game of Go, convergence to optimal performance, even though it is theoretically feasible, is not reached due to the size of the state space, which for the game of Go consists of $10^{170}$ states. 
This, however, is not worrying in practice where the goal is to create a well-performing and fast algorithm without the goal of full convergence or theoretical guarantees. 
In numerical investigations of various textbook problems, it was shown that PS demonstrates a competitive performance with respect to standard RL methods \cite{Melnikov2018,Mel15,Mau15,Mak16}. 
In this work, we complement those results by comparing PS with other RL approaches from a theoretical perspective. 
Specifically, we analyze if PS converges to an optimal solution, as other methods, like Q-learning and SARSA, have been proven to in environments which are describable by MDPs \cite{dayan1994td,singh2000convergence,jaakkola1994convergence,watkins1992q}. 
Specifically, we analyze if PS converges to an optimal solution. 
Other methods, like Q-learning and SARSA, have already been proven to converge in environments which are describable by Markov Decision Processes (MDPs) 
One should notice, however, that Q-learning and SARSA are methods equipped with update rules explicitly designed for such problems. 
PS, in contrast, was designed with a broader set of time-varying and partially observable learning environments in mind. 
For this reason, it is capable of solving tasks that a direct (naive) implementation of Q-learning and SARSA fail to learn as they are designed to obtain a time-independent optimal policy \cite{watkins1992q,bookSuttonBarto2nd}, examples can be found in \cite{Mau15}. 
Thus, it would be unlikely for a PS agent to exactly realize the same optimality with respect to the discounted infinite horizon reward figures of merit (for which Q-learning was designed) without any further adjustment to the model. 
Nonetheless, in this work, we analyze the properties of PS taken as a pure MDP solving RL algorithm. 
We show that a slightly modified PS variant recovers the notion of state-action values as a function of its internal parameters, while preserving the main characteristics that make PS stand out from other RL algorithms, such as the locality of the update rules. 
As we show, this new variant is suitable for episodic MDPs, and we can prove convergence to the optimal strategy for a range of solutions. 
In the process, we connect the modified PS model with the basic PS model, which allows us to partially explain and understand the empirical performance and successes of PS reported in previous experimental works.

This paper is organized as follows: 
We quickly recap the main concepts of RL theory \footnote{We will follow the notation introduced in \cite{bookSuttonBarto2nd} closely.} in Sec.~\ref{sect:MDP} concerning MDPs that will be used by us during the rest of this paper before we present the PS model in Sec.~\ref{sec3}.
In Sec.~\ref{sect:convergence}, we begin by introducing the adaption to PS needed for the convergence proof, which will be followed by the convergence proof that is based on a well-known theorem in stochastic approximation theory.
In the Appendix of the paper, we provide a detailed exposition of RL methods which introduces the necessary concepts for the analysis, with a broader perspective on RL theory in mind. 
Additionally, after discussing multiple variants of the PS update rules and their implications, we present an extensive investigation of the similarities and difference of PS to standard RL methods. 

\section{\label{sect:MDP} Markov Decision Processes}

\subsection{Policy and discounted return}

In the RL framework \cite{bookSuttonBarto2nd}, an \emph{RL problem} is a general concept that encompasses the learning of an agent through the interaction with an environment with the goal of maximizing some precisely defined figure of merit such as a reward function. 
In a discrete-time framework, the agent-environment interaction can be modeled as follows. At every time step $t$, the agent perceives the environmental \emph{state} $S_t$. Then the agent chooses an \emph{action} $A_t$ to execute upon the environment. 
The environment completes the cycle by signaling to the agent a new state $S_{t+1}$ and a reward $R_{t+1}$. The variables $R_t$, $S_t$ and $A_t$ are, in general, random variables, where $R_t$ can take values $r_t\in \mathbb{R}$, while $S_t$ and $A_t$ take values sampled from sets $\mathcal{S}=\{s_1,s_2,\dots\}$ and $\mathcal{A}=\{a_1,a_2,\dots\}$ respectively. 
For simplicity, we assume in the following that these two sets are finite and $r_t$ is bounded for all time steps $t$.

A particularly important set of RL problems are those where the environment satisfies the Markovian property. 
These problems can be modeled by Markov Decision Processes (MDPs). 
In an MDP, the probabilities of transitions and rewards are given by the set of probabilities
\begin{equation}
p(s',r \mid s,a)\coloneqq \Pr\{S_{t+1}=s',R_{t+1}=r \mid S_t=s,A_t=a\}.
\end{equation}

At every time step, the agent chooses an action as the result of some internal function that takes as input the current state of the environment. 
Thus, formally, an agent maps states into actions, which is captured by the so-called \emph{policy} of the agent. Mathematically, the policy (at a certain time step $t$) can be defined as the set of probabilities
\begin{equation}\label{equ.propol}
\pi(a\mid s)\coloneqq \Pr\{A_t=a\mid S_t=s\}.
\end{equation}
The successive modification of these probabilities, $\pi$ $\!=$ $\!\pi_t$,
through the experience with the environment constitutes the \emph{learning} that the agent undergoes in order to achieve a goal. In an MDP, the notion of goal can be formalized by introducing a new random variable
\begin{equation}\label{equ.Gt}
G_t(\gammadis)\coloneqq\sum_{k=0}^{\infty}\gammadis^{k}R_{t+k+1},
\end{equation}
called the \emph{discounted return}, where $\gammadis \in [0,1]$ is the \emph{discount parameter}. The case with  $\gammadis=1$ is reserved for episodic tasks, where the agent-environment interaction naturally terminates at some finite time. The discounted return at some time-step $t$ consists of the sum of all rewards received after $t$, discounted by how far in the future they are received. The \emph{solution} to the MDP is the policy that maximizes the expected return starting from any state $s$, called the \emph{optimal policy}.

A particular set of RL problems, we will consider in this work, are the so-called \emph{episodic} tasks. In these, the agent-environment interactions naturally break into episodes, e.g. an agent playing some card game, or trying to escape from a maze. Note that while in some episodic problems the objective could be to finish the episode with the fewest possible actions (e.g. escaping a maze), in general, the optimal solution is not necessarily related to ending the episode. A notion of episodic MDP can be easily incorporated into the theoretical formalism recalled above, by including a set $\mathcal{S}_T\subset\mathcal{S}$, of so-called terminal or absorbing states. These states are characterized by the fact that transitions from a terminal state lead back to the same state with unit probability and zero reward. In episodic MDPs, the goal for the agent is to maximize the expected discounted return per episode.

It should be noted that the concept of absorbing states is a theoretical construct introduced to include the concept of episodic and non-episodic MDPs into a single formalism. In a practical implementation, however, after reaching a terminal state, an agent would be reset to some \emph{initial state}, which could be a predefined state or chosen at random for instance. While such a choice could have an impact on learning rates, it is irrelevant regarding the optimal policy. For this reason, in the following, we do not make any assumption about the choice of the initial states. We will assume, however, that the environment signals the finalization of the episode to the agent.

\subsection{Value functions and optimal policy}
The concept of an optimal policy is closely intertwined with that of \emph{value functions}. 
The value $v_\pi(s)$ of a state $s\in \mathcal{S}$ under a certain policy $\pi$ is defined as the expected return after state $s$ is visited, i.e. it is the value
\begin{equation}
v_{\pi}(s)\coloneqq\E_{\pi} \left\{G_t\mid S_t=s\right\}.
\end{equation}
It has been proven for finite MDPs that there exists at least one policy, called the \emph{optimal policy} $\pi^*$, which maximizes over the space of policies $v_{\pi}(s)$ $\forall s$ simultaneously, i.e. 
\begin{equation}
v_*(s)=\max_{\pi}\left\{v_{\pi}(s)\right\},\;\forall s\in\mathcal{S},
\end{equation}
where $v_*$ denotes the value functions associated to the optimal policy.

Value functions can also be defined for state-action pairs. The so-called Q-value of a pair $(s,a)$, for a certain policy $\pi$, is defined as the expected return received by the agent following the execution of action $a$ while in state $s$, and sticking to the policy $\pi$ afterwards. The Q-values of the optimal policy, or \emph{optimal Q-values} can be written in terms of the optimal state value functions as
\begin{equation}
q_*(s,a)=r(s,a)+\gammadis \E\left\{v_*(S_{t+1})|S_t=s, A_t=a\right\},
\end{equation}
where 
\begin{equation}
r(s,a)=\E \left\{R_{t+1}\mid S_t=s,A_t=a\right\}.
\end{equation}
The relevance of Q-values is evidenced by noting that given the set of all $q_*$ values, an optimal policy can be derived straightforwardly as
\begin{equation}\label{equ.detpol}
\pi^{*}(s)=\argmax_{a'} \left\{q_*(s,a')\right\}.
\end{equation}
(Note the notational difference in the arguments to distinguish between the stochastic policy Eq.~(\ref{equ.propol}), which returns a probability, and the deterministic policy
Eq.~(\ref{equ.detpol}), which returns an action.)
For this reason, standard RL methods achieve an optimal policy in an indirect way, as a function of the internal parameters of the model, which are those which are updated through the learning of the model, and which in the limit converge to the $\qstar$ values. 
A similar procedure will be used by us in Sec.~\ref{sect:convergence}, where we discuss the convergence of PS to the optimal policy of MDPs.


\subsection{Q-learning and SARSA}
Q-learning and SARSA are two prominent algorithms that capture an essential idea of RL: online learning in an unknown environment. 
They are particularly designed to solve Markovian environments and their prominence can in part be ascribed to the theoretical results that prove their convergence in MDPs. 
In both algorithms, learning is achieved by estimating the \textit{action value function} $q_\pi(s,a)$ for every state action pair for a given policy $\pi$. 
This estimate is described be the $Q$-value which is assigned to each state-action pair. 
The update of the $Q$-value is given by:  

\begin{equation}
\begin{split}
Q_{t+1}(s_t,a_t)&= (1-\alpha)Q_{t}(s_t,a_t)+\alpha (R_{t+1}\\& +\gammadis f(Q_{t}(s_{t+1},a_{t+1})). 
\end{split}
\end{equation}
The learning rate $\alpha$ describes how fast a new estimate of the $Q$-value overwrites the previous estimate. 
In SARSA, the function $f$ is the identity, so that the $Q$-value is not only updated by the reward $R_{t+1}$ but also with the $Q$-value of the next state-action pair along the policy $\pi$. 
Thus, SARSA is an on-policy algorithm, as described in App~\ref{app:RLmethods}. 
In Q-learning, on the other hand, the function $f=\max_{a_{t+1}}$ takes the maximal $Q$-value of the next state. 
This algorithm is an off-policy algorithm due to sampling of the next action independently from the update of the $Q$-values.



\section{\label{sec3}
Projective Simulation}
Projective Simulation (PS) is a physically inspired framework for artificial intelligence introduced in \cite{Bri12}. The core of the model is a particular kind of memory called \emph{Episodic and Compositional Memory} (ECM) composed of a stochastic network of interconnected units, called clips (cf. Fig.2 in \cite{Bri12}). Clips represent either percepts or actions experienced in the past, or in more general versions of the model, combinations of those. 
The architecture of ECM, representing deliberation as a random walk in a network of clips, together with the possibility of combining clips and thus creating structures within the network, allows for modeling incipient forms of \emph{creativity} \cite{Bri12a,Han20}. Additionally, the deliberation process leading from percepts to actions has a physical interpretation in PS. Visiting any environmental state activates a corresponding percept clip in the ECM. This activation can be interpreted as an excitation, which then propagates stochastically through the network in the form of a random walk. The underlying dynamics have the potential to be implementable by real physical processes, thus relating the model to embodied agents including systems which exploit quantum effects, as has been explored in \cite{Dun15b,Cla16}.

PS can be used as an RL approach, where the action, the percept and the reward are used to update the ECM structure. 
In general, the PS framework enables to leverage complex graph structures to enhance learning. 
For example, generalization can be implemented through manipulation of the ECM topology so that the RL agent is capable of learning in scenarios it would otherwise fail to learn \cite{Mel15}. 
However, this generalization mechanism is not necessary for solving MDP environments.  

Before we discuss the ECM for solving MDPs in detail, we need to emphasis the difference between the state of the environment and the percept the agent receives. 
In an algorithm specifically designed to solve MDPs, the state contain sufficient information of the environment such that the corresponding transition function fulfills the Markov property. 
We will refer to this type of state as Markov state. 
This assumption on the state space can generally not be made in most realistic learning scenarios but it can be generalized to partially observable MDPs where the Markovian dynamics are hidden. 
In a partially observable environment, the input of the learning algorithm is an observation that is linked to a Markov state via a, from the perspective of the algorithm, unknown probability distribution. 

A percept, as introduced in the PS model, further generalizes the concept of such an observation. 
Here, the percept does not necessarily have to be connected to an underlying Markov state contrary to the observation in partially observable MDPs.
This distinction might not seem necessary for learning in a classical environment but plays a significant role when one considers quantum systems that cannot be described with hidden variable models.  
In this work, since we focus on MDPs, we will equate the percepts an agent receives and the state of the MDP.
In the following, both are denoted by $s$. 
Furthermore, we will not emphasize the difference between the percept and its corresponding percept clip, assuming there is a one-to-one correspondence between percept and percept clip.
The same holds for the actions and their corresponding action clips. 

\begin{figure}[ht]
\includegraphics[width=4cm]{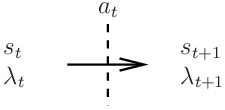}
\caption{\label{fig1}
Transition from time step $t$ to $t+1$, ($t$ $\!=$ $\!0,1,2,\ldots$), via
the agent's decision $a_t$, where $s$ and $\lambda$ denote environment state and
reward ($\lambda_0$ $\!=$ $\!0$), respectively (adapted from \cite{bookSuttonBarto2nd}).
}
\end{figure}

The ECM structure used to solve MDPs consists of one layer of percept clips that is fully connected with a layer of action clips. 
Each edge represents a state action pair $(s,a)$ which is assigned a real-valued weight (or hopping value) $h$ $\!=$ $\!h(s,a)$ and a non-negative glow value $g$ $\!=$ $\!g(s,a)$. 
While the weight $h$ determines the probability of transition between a percept clip and an action clip, the glow variable $g$ measures how ’susceptible’ this weight $h$ is to future rewards from the environment. 
In (\ref{hur}), $h^{\mathrm{eq}}$ is an (arbitrarily given) equilibrium value, and $\lambda_{t+1}$ is the reward received immediately after action $a_t$, in accordance with the time-indexing conventions in \cite{bookSuttonBarto2nd} as shown in Fig.~\ref{fig1}.

A variety of different update rules are discussed in App~\ref{app:glow} and compared with other RL methods in App~\ref{app:comp}. 
In the following, we will focus on the standard update used in \cite{Bri12,Mau15,Melnikov2018}. 
The update rules for the $h$-value and the glow value are given by:
\begin{eqnarray}
\label{hur}\nonumber
  &h_{t+1}(s,a)=h_{t}(s,a)-\gamma(h_{t}(s,a)-h^{\mathrm{eq}})\\
  &+g_{t}(s,a)\lambda_{t+1} \\ \nonumber
\label{gur1}
 & g_{t}(s,a)= (1-\delta_{(s,a),(s_t,a_t)})(1-\eta)g_{t-1}(s,a)\\
 &+\delta_{(s,a),(s_t,a_t)}
\end{eqnarray}

The update of the $h$-value consists, in the language used in the context of Master equations, of a gain and a loss term. 
The parameter for the loss term is called damping parameter and is denoted by $\gamma$ $\!\in$ $\![0,1]$. 
The parameter for the gain term is called glow parameter \cite{Mau15} and is denoted by $\eta$ $\!\in$ $\![0,1]$. 
In particular,
$\eta$ $\!=$ $\!1$ recovers the original PS as introduced in \cite{Bri12}.
Finally, $\delta_{t}$ $\!:=$ $\!\delta_{s,s_t}\delta_{a,a_t}=\delta_{(s,a),(s_t,a_t)}$ denotes the
Kronecker delta symbol, which becomes 1 if the respective $(s,a)$-pair is
visited at cycle $t$, and is otherwise 0.
The agent's policy is defined as the set of all conditional probabilities
(i.e., transition probabilities in the ECM clip network)
\begin{equation}
\label{equ:policy}
  p_{ij}=p(a_j|s_i)=\frac{\Pi({h}_{ij})}{\kappa_i},\quad
  \kappa_i=\sum_j\Pi({h}_{ij}),
\end{equation}
of selecting action $a_j$ when in state $s_i$ and is here described in terms of
some given function $\Pi$. Examples of $\Pi$ which have been used or discussed
in the context of PS are an identity function \cite{Bri12,Mau15},
\begin{equation}
\label{equ:lineal}
  \Pi(x)=x,
\end{equation}
if $x$ is non-negative, and an exponential function leading to the well-known softmax policy \cite{Melnikov2018} if a normalization factor is added
\begin{equation}
\label{equ:softmax}
  \Pi(x)=\mathrm{e}^{\beta{x}},
\end{equation}
where $\beta$ $\!\ge$ $\!0$ is a real-valued parameter.
\section{\label{sect:convergence} Convergence of PS in episodic MDPs}
In previous works, it has been shown numerically that the basic version of a PS-agent is capable of learning the optimal strategy in a variety of textbook RL problems. 
The PS model with standard update rules, however, does not necessarily converge in all MDP settings. 
This version of the PS is thoroughly analyzed in App.~\ref {app:glow} and App.~\ref{app:comp}.
As recalled in Sec.~\ref{sect:MDP}, in MDPs optimality can be defined in terms of the optimal policy.  
In this section, we present a modified version of the PS that has been designed exclusively to tackle this kind of problem. 
We consider arbitrary episodic MDPs, and derive an analytical proof of convergence. 
In this version, the policy function depends on the normalized $\htilde$ values, which, as we show, later behave similarly as state-action values, and in fact, in episodic MDPs, they converge to the optimal $\qstar$ values for a range of discount parameters.


\subsection{Projective Simulation for solving MDPs}
In the following, we introduce a new variant of PS aimed at solving episodic MDPs. 
In those problems, there is a well-defined notion of optimality, given by the optimal policy. 
As described above, the basic PS constitutes a direct policy method (see also App.~\ref{app:RLmethods}). Finding the optimal policy of an MDP by policy exploration seems a rather difficult task. 
However, as other methods have proven, finding the optimal $\qstar$ values can be done with relatively simple algorithms, and the optimal policy can be derived from the $\qstar$ values in a straightforward manner. 
Motivated by this,  we add a new local variable to the ECM-network in order to recover a notion of state-action values while maintaining the locality of the model.

For this version we consider ``first-visit'' glow, defined as follows \footnote{We can assume without loss of generality that the environment signals the finalization of the episode. Thus the first visits to an edge can be determined locally. Moreover, the same signal can be used to reset the glow values locally.}. The glow of any given edge is set to one whenever that edge is visited for the first time during an episode and in any other circumstance it is damped by a factor $(1-\eta)$, even if the same edge is visited again during the same episode. In addition, the entire glow matrix is reset to zero at the end of an episode. We thus write the updates as
\begin{eqnarray}
\label{PS1prima}
{h_{t+1}(s,a)}&=&{h_{t}(s,a)}+\lambda_{t+1}{g_{t}(s,a)}\\
\label{PS2prima}
{g_{t}(s,a)}&=&(1-\eta){g_{t-1}(s,a)}+\delta_{(s,a),(s,a)_\mathrm{first-visit}}\\ 
\label{PS3prima}
N_{t+1}(s,a)&=& N_{t}(s,a)+\delta _{(s,a),(s,a)_{\mathrm{first-visit}}} 
\end{eqnarray}
Here, the update for $h$ is the same as in Eq.~(\ref{hur}), but given that the MDPs are time-independent environments, $\gamma$ has been set to zero. 
We add a matrix $N$ to the standard PS, which counts the number of episodes during which each entry of $h$ has been updated. The idea behind these updates is that the ratios
\begin{equation}\label{PS4prima}
  \htilde_{t}(s,a)\coloneqq\frac{h_{t}(s,a)}{N+1}
\end{equation}
resemble state-action values. 
To gain some intuition about this, note that $h$-values associated to visited edges will accumulate during a single episode a sum of rewards of the form
\begin{equation}
\lambda_t+(1-\eta) \lambda_{t+1}+(1-\eta)^2 \lambda_{t+2} +\dots,
\end{equation}
which gets truncated at the time step the episode ends. Hence, the normalized $\htilde$ values become averages of sampled discounted rewards (see App~\ref{sec:eta}). Later we show that paired with the right policy and glow coefficient the $\htilde$ values converge to the optimal $\qstar$ values. 

Instead of considering a policy function of the $h$-values as in Eq.~(\ref{equ:policy}), here we will consider a policy function given by
\begin{equation}
p_{i,j} = \frac{\Pi (\tilde{h}_{i,j})}{c_i},\quad c_i=\sum_j \Pi (\tilde{h}_{i,j} ),
\end{equation}
for a certain function $\Pi(\cdot)$. Given that the $\htilde$-values are, in general, not diverging with time (in fact they are bounded in the case of bounded rewards) a linear function, as in Eq.~(\ref{equ:lineal}), would fail to converge to a deterministic policy. A solution for that is to use a softmax function as in Eq.~(\ref{equ:softmax}), where the free coefficient $\beta$ is made time-dependent. By letting $\beta$ diverge with time, the policy can become deterministic in the limit.

Similarly to Monte Carlo methods, which may be equipped with a variety of update rules, giving rise to first-visit or many-visit Monte Carlo methods, the choice of the glow update rule is to some extent arbitrary but may depend on the physical implementation of PS and the ECM. For example, instead of Eq.~(\ref{PS1prima}), one could use the accumulating glow update, given in Eq.~(\ref{gur2}). In that case, one simply needs to change the update rule of $N$, given in Eq.~(\ref{PS3prima}) in such a way that every visit of the edge is counted, instead of only first visits. Intuitively, both pairs of update rules have similar effects, in the sense that in both cases $\htilde(s,a)$ equals an average of sampled discounted returns starting from the time a state-action pair $(s,a)$ was visited. However, while for first-visit glow we were able to prove convergence, that is not the case for accumulating glow. Therefore, when referring to this version of PS in the following, we assume update rules given by (\ref{PS1prima})-(\ref{PS3prima}).

\subsection{Convergence to the optimal policy}

The convergence of $\htilde$ values to $\qstar$ values can be proven by a standard approach used in the literature to prove, for example, the convergence of RL methods like Q-Learning and SARSA, or prediction methods like TD($\lambda$). In the remainder of the paper, we will use interchangeably the boldface notation $\edge$ to denote a state-action pair as well as the explicit notation
$(s,a)$ whenever convenience dictates.
Denoting by $\htilde_m (\edge)$ the $\htilde$-value of edge $\edge$ at the end of episode $m$, we define the family of random variables $\Delta_m (\edge) \coloneqq \htilde_m (\edge) - \qstar(\edge)$. We want to show that in the limit of large $m$, $\Delta _m (\edge)$ converges to zero for all $\edge$. Moreover, it is desirable that such convergence occurs in a strong sense, i.e. with probability one. We show that by following the standard approach of constructing an update rule for $\Delta_m (\edge)$ which satisfies the conditions of the following theorem \footnote{ This theorem is a known result in the field of Stochastic Approximation. While the first version of the theorem was presented in \cite{ dvoretzky1956stochastic}, it can be found in many forms in the literature. The version presented here, where the contraction property has been relaxed by allowing a noise that tends to zero, has been presented in \cite{singh2000convergence}.}

\begin{theorem}\label{theorem}
	
	A random iterative process $\Delta_{m+1} (\bm{x}) =\left[1-\alpha_m (\bm{x})\right]\Delta_m (\bm{x}) +\alpha_m (\bm{x}) F_m (\bm{x})$, $\bm{x}\in X$ converges to zero with probability one (w.p.1) if the following properties hold:
	\begin{enumerate}
		
		\item the set of possible states $X$ is finite. 
		
		\item $0\leq \alpha_m (\bm{x}) \leq 1$, $\sum_m \alpha_m (\bm{x}) = \infty $, $\sum_m \alpha_m ^2 (\bm{x}) < \infty $ w.p.1, where the probability is over the learning rates $\alpha_m (\bm{x}) $.
		
		\item $\normw{\E\{F_m(\cdot)|P_m\}}\leq \kappa \normw{\Delta_m (\cdot ) }+c_m$, where $\kappa \in [0,1)$ and $c_m$ converges to zero w.p.1
		
		\item $\operatorname{Var}\{F_m (\bm{x}) | P_m\}\leq K (1+\normw{\Delta_m (\cdot) })^2$, where $K$  is some constant.
	\end{enumerate}
	Here $P_m$  is the past of the process at step $m$, and the notation $\normw{\cdot}$ denotes some fixed weighted maximum norm. 
	
\end{theorem}

In addition to $\Delta_m(\edge)$ meeting the conditions of the theorem, the policy function must also satisfy two specific requirements. 
First of all, it must be \emph{greedy} with respect to the $\htilde$-values (at least in the limit of $m$ to infinity). In that way, provided that the $\htilde$-values converge to the optimal $\qstar$ values, the policy becomes automatically an optimal policy. Additionally, to guarantee that all $\Delta_m$ keep being periodically updated, the policy must guarantee infinite exploration. A policy that satisfies these two properties is called GLIE \cite{singh2000convergence}, standing for Greedy in the Limit and Infinite Exploration. Adapting the results from \cite{singh2000convergence} for PS and episodic environments, we can show
(see App.~\ref{sect:app_3}) that a softmax policy function defined by
\begin{equation}
\pi_m(a|s,\bm{\htilde}_m)=\frac{\exp\left[\beta_m\tilde{h}_m(s,a)\right]}{\sum_{a'\in\mathcal{A}}\exp\left[\beta_m\tilde{h}_m(s,a')	\right]}
\end{equation}
is GLIE, provided that $\beta_m\rightarrow_{m\rightarrow \infty}\infty$ and $\beta_m\leq C \ln(m)$, where $C$ is a constant depending on $\eta$ and $\left|\mathcal{S}\right|$. While the first condition on $\beta_m$ guarantees that the policy is greedy in the limit, the second one guarantees that the agent will keep exploring all state-action pairs infinitely often. In this particular example, we have considered $\beta_m$ to depend exclusively on the episode index. By doing so, the policy remains local, because $\beta_m$ can be updated using exclusively the signal of the environment indicating the finalization of the episode. Note however that the choice of the policy function, as far as it is GLIE, has no impact on the convergence proof.
We are now in a position to state our main result about the convergence of PS-agents in the form of the following theorem. 

\begin{theorem}\label{lemma}
	
	For any finite episodic MDP with a discount factor of $\gammadis$, the policy resulting from the new updates converges with probability one to the optimal policy, provided that
	\begin{enumerate}
		\item rewards are bounded,
		\item $0\leq \gamma_{\mathrm{dis}} \leq 1/3 $ ,
                      where $\gammadis=1-\eta$,
		\item the policy is a GLIE function of the $\htilde$-values.
	\end{enumerate}	
\end{theorem}

Note that we have restricted the range of values $\gammadis$ can take. 
The reason for that is related to the way the $h$-values are updated in PS. 
In Q-Learning and SARSA, where the $\gammadis$ parameter of the MDP is directly included in the algorithm, every time an action is taken its corresponding Q-value is updated by a sum of a single reward and a discounted bootstrapping term. 
Given that the PS updates do not use bootstrapping, that term is ``replaced'' by a discounted sum of rewards. 
Due to this difference, the contraction property (Condition 3 in Theorem \ref{theorem}) is not so straightforward to prove forcing us to consider smaller values of $\gammadis$.
However, this condition on the $\gammadis$ parameter is not a fundamental restriction of the PS model, but merely a result of how convergence is proven in this work. 

\subsection{Environments without terminal states}

In Theorem \ref{lemma}, we have considered exclusively episodic MDPs. 
However, it is still possible for these environments to have an optimal policy which does not drive the agent to any terminal states. This observation suggests that the scope of problems solvable by PS can be further extended to a subset of non-episodic MDPs.

Given any non-episodic MDP, one can construct an episodic MDP from it by adding one single terminal state $s_T$ and one single transition leading to it with non-zero probability, i.e. by defining $p_T=\Pr(s_T|s,a)\neq 0$ for some arbitrary pair $(s,a)$. Thus, while the original non-episodic MDP falls outside the scope of Theorem \ref{lemma}, PS could be used to tackle the non-episodic MDP. Anyway, in general, these two problems might have different solutions, i.e. different optimal policies. However, given that both the pair $(s,a)$ for which $p_T\neq 0$ and the value of $p_T$ are arbitrary, by properly choosing them, the difference between the two optimal policies could become negligible or non-existent. That could be done easily having some useful knowledge about the solution of the original MDP. Consider for instance a grid world, where multiple rewards are placed randomly around some central area of grid cells. Even without knowing the exact optimal policy, one can correctly guess that there will be an optimal cyclic path about the center of the world yielding the maximum expected discounted return. Hence, adding a terminal state in some remote corner of the world would very likely leave the optimal policy unchanged.


\subsection{\label{sect:proof} Proof of Theorem \ref{lemma}}

In this section, we discuss the core of the proof of Theorem \ref{lemma}, leaving for App.~\ref{app:details} the most involved calculations. Given that the policy is a greedy-in-the-limit-function of the $\htilde_m$ values, the proof of Theorem \ref{lemma} follows if we show that
\begin{equation}\label{equ.deltadefinition}
\Delta_m (\edge) \coloneqq \tilde{h}_m (\edge) -\qstar (\edge)
\end{equation} 
converges to zero with probability one. In order to do so, we show that $\Delta_m(\edge)$ obeys an update rule of the form given in Theorem \ref{theorem} and the four conditions of the theorem are satisfied.

We begin by deriving an update rule for the $h$-values between episodes. 
In the case where an edge $\edge$ is not visited during the $m$-th episode, its corresponding $h$-value is left unchanged, i.e. $h_m(\edge)=h_{m-1}(\edge)$. 
Otherwise, due to the decreasing value of the glow during the episode, in the $m$-th episode, the $h(\edge)$ value will accumulate a discounted sum of rewards given by 
\begin{equation}
D_m (\edge) =\sum_{t=t_m(\edge)} ^{T_m} \etabar ^{t-t_m(\edge)} \lambda_t,
\end{equation}
where $t_m (\edge) $ and $T_m$ are the times at which the first visit to $\edge$ during episode $m$ occurred and at which the episode finished, respectively, and $\etabar=1-\eta$.
Therefore, in general $h_m(\edge)=h_{m-1}(\edge)+\chi_m(\edge)D_m(\edge)$, where $\chi_m(\edge)$ is given by 

\begin{equation}
\chi_m(\edge)=
\begin{cases}
1\quad\textrm{if $\edge$ is visited during the $m$-th episode,}\\
0\quad\textrm{otherwise.}\\
\end{cases}
\end{equation}

We denote, respectively, by $N_m(\edge)$ and $\htilde_m (\edge)$ the $N$-value and $\htilde$-value associated to edge $\edge$ at the end of episode $m$. Thus, we have that $ \htilde_m (\edge)= h_m(\edge)/[N_m(\edge)+1]$ and it obeys an update rule of the form
\begin{equation}\label{equ.htildeupdate}
\begin{aligned}
\tilde{h}_m (\edge) =\frac{1}{N_m(\edge)+1}\Big\{&\big[N_{m-1}(\edge)+1\big] \tilde{h}_{m-1} (\edge)\\
&+\chi_m(\edge)D_m (\edge)\Big\}.
\end{aligned}
\end{equation}
Noting that the variables $N_m(\edge)$ can be written in terms of $\chi_m(\edge)$ as the sum $N_m(\edge)=\sum_{j=1}^{m}\chi_j (\edge)$, it follows from Eq.~(\ref{equ.htildeupdate}) that the variable $\Delta_m(\edge)$ given in Eq.~(\ref{equ.deltadefinition}) satisfies the recursive relation
\begin{equation}\label{equ.deltaupdate}
\Delta_{m} (\edge) =\left[1-\alpha_m (\edge)\right]\Delta_{m-1} (\edge) +\alpha_m (\edge) F_m (\edge),
\end{equation}
where the ratios
\begin{equation}\label{equ.alpha_m}
\alpha_m (\edge)\coloneqq\frac{\chi_m (\edge)} {N_m (\edge)+1}, 
\end{equation}
play the role of learning rates, and $F_m(\edge)$ is defined as
\begin{equation}\label{eq:F_definition}
F_m (\edge)\coloneqq \chi_m (\edge) \left(D_m(\edge) -\qstar(\edge) \right).
\end{equation}
The update rule in Eq.~(\ref{equ.deltaupdate}) is exactly of the form given in Theorem \ref{theorem}. Therefore we are left with showing that $\alpha_m(\edge)$ satisfies Condition 2 in Theorem \ref{theorem}, and  $F_m(\edge)$ satisfies Conditions 3 and 4. Below we describe the general procedure to prove that, while most of the details can be found in App.~\ref{app:details}. 

The fact that $\alpha_m(\edge)$ satisfies Condition 2 in Theorem \ref{theorem} follows from noting that $\sum_m \alpha_m (\edge)=\sum_n 1/n$ and $\sum_n \alpha^2_m(\edge)=\sum_n 1/n^2$, which are, respectively, a divergent and a convergent series. Regarding Condition 3, note that by tweaking the free glow parameter in such a way that $\etabar = \gamma_{\mathrm{dis}}$,
the variable $D_m (\edge)$ becomes a truncated sample of the discounted return $G(\edge,\gammadis)$ given in Eq.~(\ref{equ.Gt}). Thus, $\htilde$ values undergo a similar update to that found in SARSA, with the difference that instead of a bootstrapping term an actual sample of rewards is used. Due to these similarities we can use the same techniques used in the proof of convergence of RL methods \cite{jaakkola1994convergence,singh2000convergence} and show that
\begin{equation}\label{eq:S_jk_inequality}
\normw{\E\{F_m(\cdot)|P_m\}}\leq f(\gammadis) \normw{\Delta_m (\cdot)}+c_m,
\end{equation}
where $c_m$ converges to zero w.p.1 and $f(\gammadis)=\frac{2\gammadis}{1-\gammadis} $.
This equation satisfies Condition 3 in Theorem \ref{theorem} as far as $f(\gammadis)<1$, which occurs for $\gammadis<1/3$. 

Finally, Condition 4 in Theorem \ref{theorem} follows from the fact that rewards are bounded. This implies that $\htilde$-values and, in turn, the variance of $F_m(\edge)$ are bounded as well. This concludes the proof of Theorem \ref{lemma}.

\section{\label{sect:conclusion} Conclusion}

In this work, we studied the convergence of a variant of PS applied to episodic MDPs. Given that MDPs have a clear definition of a goal, characterized by the optimal policy, we took the approach of adapting the PS model to deal with this kind of problem specifically. The first visit glow-version of PS presented in this work, internally recovers a certain notion of state-action values, while preserving the locality of the parameter updates, crucial to guarantee a physical implementation of the model by simple means. We have shown that with this model a PS-agent achieves optimal behavior in episodic MDPs, for a range of discount parameters. 
This proof and the theoretical analysis of the PS update rules shed light on how PS, or, more precisely, its policy, behaves in a general RL problem.

The PS updates that alter the h-values at every time step asynchronously pose a particular challenge for proving convergence. 
To deal with that, we analyzed the subsequence of internal parameters at the times when episodes end, thus recovering a synchronous update. 
We could then apply techniques from stochastic approximation theory to prove that the internal parameters of PS converge to the optimal $q$ values, similarly as in the convergence proofs of other RL methods.

We have also chosen a specific glow update rule, which we have called first-visit glow. While other glow updates, like accumulating or replacing glow, show the same behavior at an intuitive level, trying to prove the convergence with those updates has proven to be more cumbersome. Therefore, from a practical point of view, several glow mechanisms could be potentially utilized, but convergence in the limit is, at the moment, only guaranteed for first-visit glow.

Although only episodic MDPs fall within the scope of our theorem, no constraints are imposed on the nature of the optimal policy. Hence, episodic problems where the optimal policy completely avoids terminal states (i.e. the probability that an agent reaches a terminal state by following that policy is strictly zero) can also be considered. Furthermore, the agent could be equipped with any policy, as far as the GLIE condition is satisfied. In this paper, we provided a particular example of a GLIE policy function, in the form of a softmax function with a global parameter, which depends exclusively on the episode index. In this particular case, the policy is compatible with local updates, in the sense that the probabilities to take an action given a state can be computed locally.

\section{Acknowledgements}
This work was supported, in part, by the Austrian Science Fund through the projects SFB FoQus F4212, SFB BeyondC F7102, and DK ALM W1259, and in part by the  Dutch Research Council (NWO/OCW), through the Quantum Software Consortium programme (project number  024.003.037).
\appendix


\section{\label{app:RLmethods}
A review of RL methods}
The following section is meant as a concise overview of standard RL methods, which we distilled and adapted from \cite{bookSuttonBarto2nd}, to provide the necessary background before which the PS will we be discussed in Sections \ref{app:glow} and \ref{app:comp}. For details we refer the reader to Ref. \cite{bookSuttonBarto2nd}.

\begin{figure}[ht]
\includegraphics[width=8cm]{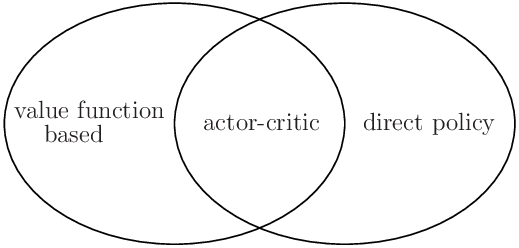}
\caption{\label{fig3}
Some types of gradient-based solution methods for RL-problems.
Value function-based methods are gradient-descent with respect to a TD-error,
whereas direct policy methods are gradient-ascent with respect to the expected
return. Actor-critic methods are depicted as their intersection since
they combine both approaches. Being understood as ``parametric'' methods, this
figure corresponds to the left branch of Fig. 3 in \cite{Cla16}.
}
\end{figure}

Among the model-free and gradient-based approaches we can broadly distinguish
between value function-based methods which are gradient-descent with respect to
a so-called temporal difference (TD) error and direct policy methods which are
gradient-ascent with respect to the expected return as shown in Fig.~\ref{fig3}.

In what follows, we focus on actor-critic methods because they exhibit an
``all in one'' structure from which the other approaches can be deduced by
simplifications. To make it short and provide an overall picture, the so-called
update rules for a single time step are listed in (\ref{acs}) and will be
explained in the remainder of this section.
\begin{subequations}
\label{acs}
\begin{eqnarray}
\label{ac-a}
  \delta&\leftarrow&R+\gammadis\,{u}^\prime-{u}\hspace{1.1cm}\mathrm{(episodic)},
  \\
\label{ac-b}
  \delta&\leftarrow&R-\bar{R}+{u}^\prime-{u}\hspace{0.6cm}\mathrm{(continuing)},
  \\&&\nonumber\\
\label{ac-c}
  \bm{z}^{u}
  &\leftarrow&\gammadis\,\lambda_\mathrm{tra}^{u}\,\bm{z}^{u}+\Gamma\,\bm{\nabla}{u},
  \\
\label{ac-d}
  \bm{\theta}^{u}
  &\leftarrow&\bm{\theta}^{u}+\alpha^{u}\,\delta\,\bm{z}^{u},
  \\&&\nonumber\\
\label{ac-e}
  \bm{z}^{\pi}
  &\leftarrow&\gammadis\,\lambda_\mathrm{tra}^{\pi}\,\bm{z}^{\pi}+\Gamma\,\bm{\nabla}\ln{\pi},
  \\
\label{ac-f}
  \bm{\theta}^{\pi}
  &\leftarrow&\bm{\theta}^{\pi}+\alpha^{\pi}\,\delta\,\bm{z}^{\pi},
  \\&&\nonumber\\
\label{ac-g}
  \Gamma&\leftarrow&\gammadis\,\Gamma\hspace{2.5cm}\mathrm{(episodic)},
  \\
\label{ac-h}
  \bar{R}&\leftarrow&\bar{R}+\eta_\text{}\delta\hspace{1.9cm}\mathrm{(continuing)}.
\end{eqnarray}
\end{subequations}

In (\ref{acs}), we have two players: an actor [the policy
$\pi$ $\!=$ $\!\pi(A|S,\bm{\theta}^{\pi})$] and a critic [the value function
$u$ $\!=$ $\!u(S,\bm{\theta}^{u})$], hence all corresponding quantities are
distinguished by their respective superscript.
The value function is parameterized by a weight vector $\bm{\theta}$. 
The vector $\bm{z}$ is referred to as the eligibility trace. 

The update equations for the actor are given by (\ref{ac-e})--(\ref{ac-f}) and the updates for the critic are  (\ref{ac-c})--(\ref{ac-d}), where the $\bm{\nabla}$ $\!:=$ $\!\frac{\partial}{\partial\bm{\theta}}$
denote the gradients with respect to the $\bm{\theta}$ vectors. These two sets of updates are identical except for the
natural logarithm $\ln{\pi}$ of the policy taken in (\ref{ac-e}). This logarithm
is a consequence of the policy gradient theorem and makes the gradient of the
actual performance measure to be maximized (value of the start state of an
episode or average rate of reward per time step for continuing problems, see
below) independent of the derivative of the state distribution (that is, the
effect of the policy on the state distribution), which depends on the
environment and is unknown.

Eqs.~(\ref{acs}) describe \emph{approximative} methods, since they apply 
\textbf{function approximation} as a scalable way of generalizing from a state
space much larger then the memory capacity of the agent. The \textbf{tabular}
case can be recovered from this as a special \emph{exact} case, in the sense
that all encountered states and actions are represented individually. The
function approximations are hidden in the gradients $\bm{\nabla}{u}$ and
$\bm{\nabla}\ln{\pi}$ in (\ref{ac-c}) and (\ref{ac-e}) and can be done in
linear fashion [by scalar products $\bm{\theta}\bm{\cdot}\bm{x}$ with feature basis
vectors $\bm{x}(S)$ or $\bm{x}(S,A)$] or in nonlinear fashion
(e.g. by neural networks with $\bm{\theta}$ as connection weights)\footnote{These examples are not exhaustive. In a wider sense, one may also
mention decision trees with $\bm{\theta}$ defining the split points and leaf
values.}.
Note that the two parametrizations $\bm{\theta}^{u}$ and
$\bm{\theta}^{\pi}$ are entirely unrelated (and hence different-dimensional in
general). In App.~\ref{app:A0}, we show in the example of SARSA, how tabular
methods can be recovered from (\ref{acs}).

In (\ref{acs}), we can use a state value function,
$u$ $\!=$ $\!u(S,\bm{\theta}^{u})$, because the policy is taken care of
separately. Without it, i.e., when we only know the value $u(s)$ of the state
$s$ we are in, we would require an \textbf{environment model}
$p(s^\prime,r|s,a)$ to decide on an action $a$. To remain model-free, we would
then have to apply an action value function
$u$ $\!=$ $\!u(s,a,\bm{\theta}^{u})$ instead, from which could obtain the best
action by search for argmax$_au$.

Eqs.~(\ref{acs}) contain six meta parameters: $\eta$ $\!>$ $\!0$ and the two
$\alpha$ $\!>$ $\!0$ are step sizes, $\gammadis$ $\!\in$ $\![0,1]$ is the
discount-rate parameter, and the two $\lambda_\mathrm{tra}$ $\!\in$ $\![0,1]$ are trace-decay
rates that allow to vary the degree of \textbf{bootstrapping}, which denotes the
updating of estimates by using other existing estimates
(cf. App.~\ref{app:A-1}). In (\ref{acs}), these existing estimates involve the
current values ${u}^\prime$ of subsequent (i.e., one time step later) states or
state-action pairs, which enter the TD-error $\delta$ in either (\ref{ac-a}) or
(\ref{ac-b}) together with the reward $R$. Choosing $\lambda_\mathrm{tra}$ is thus a
possibility to interpolate between the fully bootstrapping original one-step
TD methods which are recovered for $\lambda_\mathrm{tra}$ $\!=$ $\!0$, and
\textbf{Monte Carlo} (i.e., non-bootstrapping) methods, which are obtained in
the limit $\lambda_\mathrm{tra}$ $\!=$ $\!1$. Monte Carlo methods rely exclusively on actual
complete returns $G_t$ received. In a strict sense, they update
\textbf{off-line}, i.e., they store a whole episode
$S_0,$ $\!A_0,$ $\!R_1,$ $\!\ldots,$ $\!S_{T-1},$ $\!A_{T-1},$ $\!R_{T},$
$\!S_{T}$ in a separate memory and only at the end of an episode the respective
current estimates are updated in reverse time order
$t$ $\!=$ $\!T-1,$ $\!T-2,$ $\!\ldots,$ $\!0$ making use of the fact that
${G}_{t}$ $\!=$ $\!R_{t+1}+\gammadis{G}_{t+1}$. In contrast to
the updates (\ref{acs}), which are are done \textbf{online}
(i.e., are incremental step-by-step), strict Monte Carlo methods are thus
incremental in an episode-by-episode sense, and are consequently only defined
for the episodic case. Consequently, even for $\lambda_\mathrm{tra}$ $\!=$ $\!1$
the online updates (\ref{acs}) approximate Monte Carlo methods only for
infinitesimally small step sizes $\alpha$.

In continuing problems, the interaction between agent and environment goes
on forever without termination or start states. Discounting is here useful in
the tabular case but problematic when used with function approximation, where
the states cannot be clearly distinguished anymore. An alternative then is to
use the average rate of reward
$r$ $\!:=$ $\!\lim_{T\to\infty}\frac{1}{T}\sum_{t=1}^TE(R_t)$ $\!=$
$\!\lim_{t\to\infty}E(R_t)$, i.e., the average reward per time step
(assuming ergodicity). $E(R_t)$ is the respective expected reward and
(\ref{equ.Gt}) is replaced with the differential return
$G_t$ $\!:=$ $\!\sum_{k=0}^{\infty}(R_{t+k+1}-r)$ . In
(\ref{acs}), we thus set $\gammadis$ $\!=$ $\!1$ in such a case and apply
(\ref{ac-b}) and (\ref{ac-h}) instead of (\ref{ac-a}) and (\ref{ac-g}),
respectively. $\bar{R}$ is the current estimate of the average reward $r$.

To actually run (\ref{acs}), $\bar{R}$ and $\bm{\theta}$ can be initialized
arbitrarily (e.g. to 0). At the beginning of each episode, $\bm{z}$ is
initialized to 0, and $\Gamma$ is initialized to 1. At the end of each episode,
the value of a terminal state is set to 0.

Eqs.~(\ref{acs}) are \textbf{on-policy} and must be distinguished from
\textbf{off-policy} methods (such as Q-learning) which train on a distribution
of transitions that is different from the one corresponding to the targeted
(desired) behavior and thus free what the agent is actually doing (behavior
policy) from what it should do (target policy). While this is not fundamentally
required for basic problems that can be treated with model-free learning, it
becomes essential in (model-based) prediction and planning, where it allows
parallel learning of many target policies from the (necessarily one) realized
behavior policy. 
Combining function approximation, bootstrapping, and off-policy updates may lead to instability and divergence.

At first glance, (\ref{acs}) look elaborate and one might wonder why, for instance, value function-based methods should not suffice.
The short answer is that this depends on the type and intricacy of the problem.
To be more specific, one reason is that the expected return of state-action-pairs, that value functions estimate, typically contains more information than needed to make decisions. 
For example, a transformation of $u$ which leaves the order of its values unchanged (such as multiplication with a positive constant) has no effect on the respective optimal decision. 
As a consequence, value function-based methods are too strict, since $u$ are well defined and one needs to separately decide
on a policy in order to convert these value estimates to action selection
probabilities. If, for example a so-called softmax policy (\ref{equ:softmax})
is used, a choice and schedule (i.e., time-dependence) of the so-called inverse
temperature parameter $\beta$ has to be made. In contrast, direct policy methods, for instance, internally work with numerical preferences
$h(s,a,\bm{\theta}^{\pi})$ whose actual values do not have to represent anything
meaningful and are thus free to evolve in parameter space.
\section{\label{app:A0}
Recovering SARSA from actor-critic methods}
To recover a pure action value method from the actor-critic methods (\ref{acs}),
we restrict attention to (\ref{ac-a}), (\ref{ac-c}), and (\ref{ac-d}), set
$\Gamma$ $\!=$ $\!1$ and ignore the remaining updates. For $u$ we choose an
action value function $u$ $\!=$ $\!u(S,A,\bm{\theta}^{u})$ which we name as $q$.
Suppressing the superscript $u$ but adding time step indices, this gives for the
scalar TD-error signal 
\begin{equation}
\label{TDe}
 \delta^{\mathrm{TD}}_t=R_{t+1}+\gammadis{q}_t(S_{t+1},A_{t+1})-{q}_t(S_t,A_t),
\end{equation}
with which the remaining updates describe SARSA($\lambda_\mathrm{tra}$) with function
approximation,
\begin{eqnarray}
\label{uqf1}
  \bm{\theta}_{t+1}&=&\bm{\theta}_{t}+\alpha\delta^{\mathrm{TD}}_t\bm{z}_{t},
  \\
\label{uqf2}
  \bm{z}_{t+1}&=&\gammadis \lambda_\mathrm{tra}\bm{z}_{t}+\bm{\nabla}{q}_{t+1}(S_{t+1},A_{t+1}).
\end{eqnarray}
In the tabular case, $q$ becomes a table (matrix) with entries $q(S_i,A_j)$.
The components of the parameter vector $\bm{\theta}$ are identified with just
these entries, so that the gradient becomes a table of Kronecker delta symbols
\begin{equation}
\label{Kmatrix}
  \bm{\nabla}{q}(S,A)|_t=\frac{\partial{q}(S,A)}{\partial{q}(S_t,A_t)}
  =(\delta_{S,S_t}\delta_{A,A_t})=:\bm{\delta}_{t}.
\end{equation}
To clarify, the bold $\bm{\delta}_{t}$ has the same dimension as $q$ (i.e., it
is a matrix) with the single entry corresponding to $(S_t,A_t)$ 
(i.e., the state-action-pair visited at time $t$) being equal to 1
and all other entries being 0 and must be distinguished from the non-bold
$\delta_{t}$ used throughout Sec.~\ref{sec3}, which refers to a single given
state-action-pair, and is 1 (0), if this pair is (not) visited at time
$t$. With the bold $\bm{\delta}_{t}$, the updates (\ref{uqf1})--(\ref{uqf2})
reduce to tabular SARSA($\lambda_\mathrm{tra}$),
\begin{eqnarray}
\label{uq1}
  {q}_{t+1}&=&{q}_{t}+\alpha\delta^{\mathrm{TD}}_t\bm{z}_{t},
  \\
\label{uq2}
  \bm{z}_{t+1}&=&\gammadis\lambda_\mathrm{tra}\bm{z}_{t}+\bm{\delta}_{t+1},
\end{eqnarray}
where $\bm{z}$ is here called an accumulating eligibility trace and also
has the same dimension as ${q}$ (i.e., it is also a matrix). Hence (\ref{uq1})
updates \emph{all} entries of ${q}$. For $\lambda_\mathrm{tra}$ $\!=$ $\!0$, only the
respective (i.e., visited) \emph{single} entry of ${q}$ is updated,
\begin{equation}
  {q}_{t+1}(S_t,A_t)={q}_{t}(S_t,A_t)+\alpha\delta^{\mathrm{TD}}_t,
\end{equation}
which corresponds to conventional one-step SARSA.

\section{\label{app:A-1}
Notes on eligibility trace vectors}
We can here only outline a sketch. Let us focus on value function methods, where
for simplicity of notation we restrict attention to state values. Action value
methods can be covered analogously by referring to $(s,a)$-pairs instead of
states $s$. We may consider the mean squared value error
\begin{equation}
  \overline{VE}(\bm{\theta})
  :=\sum_{s}\mu(s)\left[u(s)-\hat{v}(s,\bm{\theta})\right]^2
\end{equation}
between the true value function $u(s)$ and a current estimate
$\hat{v}(s,\bm{\theta})$ of it.
$\mu(s)$ can be any probability distribution, but is typically chosen to be
the fraction of time spent in $s$ under on-policy training in which case it is
called ``on-policy distribution'' (or stationary distribution in continuing
tasks). Ideally, one would find a global optimum $\bm{\theta}^*$ such that
$\overline{VE}(\bm{\theta}^*)$ $\!\le$ $\!\overline{VE}(\bm{\theta})$
$\forall\bm{\theta}$. The problem is that $u(S_t)$ is unknown, hence we
substitute a so-called update target (``backed-up'' value) $U_t$ as a random
approximation of the true value $u(S_t)$ and apply stochastic gradient descent
\begin{eqnarray}
\label{ia0}
  \bm{\theta}_{t+1}&=&\bm{\theta}_{t}-\frac{1}{2}\alpha_t\bm{\nabla}
  \left[U_t-\hat{v}(S_t,\bm{\theta}_t)\right]^2
  \\
\label{ia}
  &=&\bm{\theta}_{t}+\alpha_t\left[U_t-\hat{v}(S_t,\bm{\theta}_t)\right]
  \bm{\nabla}\hat{v}(S_t,\bm{\theta}_t).
\end{eqnarray}
If $U_t$ is an unbiased estimate of $u(S_t)$, i.e.,
$\mathbb{E}\left[U_t|S_t=s\right]$ $\!=$ $\!u(s)$ $\forall$ $\!t$, then
convergence to a local optimum (i.e., the above inequality holds in a
neighborhood of $\bm{\theta}^*$) follows under the stochastic approximation
conditions for the step-size parameter $\alpha_t$ $\!>$ $\!0$,
\begin{equation}
\label{cc}
  \sum_{t=0}^{\infty}\alpha_t=\infty,\quad\sum_{t=0}^{\infty}\alpha_t^2<\infty.
\end{equation}

One possible choice for $U_t$ is the $\lambda$-return
\begin{equation}
\label{lambdareturn}
  G_{t}^{\lambda}:=(1-\lambda)\sum_{n=1}^{\infty}\lambda_\mathrm{tra}^{n-1}G_{t:t+n}
\end{equation}
as a mixture ($\lambda_\mathrm{tra}$ $\!\in$ $\![0,1]$) of $n$-step returns at time $t$,
\begin{equation}
\label{nstepreturn}
  G_{t:t+n}:=\sum_{k=0}^{n-1}\gammadis^kR_{t+k+1}
  +\gamma^n\hat{v}(S_{t+n},\bm{\theta}_{t+n-1}).
\end{equation}
Referring to episodic tasks with (random) termination time $T$, i.e.,
$t$ $\le$ $T$ $\!-$ $\!n$ in (\ref{nstepreturn}), and
$G_{t:t+n}$ $\!=$ $\!G_{t:T}$ $\!=$ $\!G_{t}$ for $n$ $\ge$ $T$ $\!-$ $\!t$,
one can decompose
\begin{equation}
  G_{t}^{\lambda_\mathrm{tra}}=(1-\lambda_\mathrm{tra})\sum_{n=1}^{T-t-1}\lambda_\mathrm{tra}^{n-1}G_{t:t+n}
  +\lambda_\mathrm{tra}^{T-t-1}G_{t},
\end{equation}
in order to demonstrate the TD(0)-limit
$G_{t}^{\lambda}$ $\!\stackrel{\lambda_\mathrm{tra}\to0}{\to}$ $\!G_{t:t+1}$ of recovering
the  one-step return and the Monte Carlo limit
$G_{t}^{\lambda}$ $\!\stackrel{\lambda_\mathrm{tra}\to1}{\to}$ $\!G_{t}$
of recovering the conventional (full) return (\ref{equ.Gt}).

The incremental updates (\ref{ia}) refer to time $t$  but involve via
(\ref{nstepreturn}) information that is only obtained after $t$. Thus we must
rearrange these updates such that they are postponed to later times following a
state visit. This is accomplished by eligibility trace vector updates such as
(\ref{ac-c}). To see this, one may sum up all increments (\ref{ac-d}) over an
episode [with (\ref{ac-c}) substituted in explicit form] and compare this with
the sum of all increments (\ref{ia}) [with $U_t$ $\!=$ $\!G_{t}^{\lambda}$] over
an episode. Both total increments are equal if we neglect the influence of the
change of $\bm{\theta}$ on the target $U_t$. This approximation holds for
sufficiently small $\alpha_t$, but strictly speaking, the bootstrapping involved
in $G_{t}^{\lambda}$, namely its dependence on $\bm{\theta}$ via
$\hat{v}(S_{t+n},\bm{\theta}_{t+n-1})$ renders the transition from (\ref{ia0})
to (\ref{ia}) only a (``semi-gradient'') approximation.

While a Monte Carlo target $U_t$ $\!=$ $\!G_{t}$ is by definition unbiased,
Monte Carlo methods are plagued with high variance and hence slow convergence.
In contrast, bootstrapping reduces memory resources, is more data-efficient, and
often results in faster learning, but introduces bias. This is reminiscent of
the ``bias-variance tradeoff'' which originally refers to supervised learning.
While in the latter case the ``trading'' is realized by methods such as
``regularization'' or ``boosting'' and ``bagging'', in the context of RL
considered here, choosing $\lambda_\mathrm{tra}$ serves a similar function.

\section{Choice of glow update}
\label{app:glow}
In the following, different types of glow updates are discussed, which are useful for the comparison with other RL methods in App.~\ref{app:comp}. 
We will focus on two types of glow, which we refer to as replacing and accumulating glow,
motivated by the respective eligibility traces of the same name, which were
originally introduced as tabular versions of the vector $\bm{z}$ and represent
in the methods discussed in App.~\ref{app:RLmethods} the counterpart of glow introduced
in Sec.~\ref{sec3}.
While replacing glow defined in Equation (\ref{gur1}) is the version applied in all works
on PS so far, accumulating glow defined in Equation (\ref{gur2}) and first-visit glow
defined in Equation (\ref{PS1prima}) are introduced to simplify the expressions and
analysis. From the perspective of the methods considered in
App.~\ref{app:RLmethods}, accumulating glow  is more similar to the $\bm{z}$-updates
(\ref{ac-c}) and (\ref{ac-e}) than replacing glow, for which such a
generalization is not clear. As far as the choice of an eligibility trace in
tabular RL methods is concerned, the tabular case of (\ref{acs}) yields
accumulating traces as is shown in (\ref{uq1})--(\ref{uq2}) in the example of
SARSA. 

\subsection{Replacing glow}
In what follows, we discuss the update rules for replacing glow, which is helpful for the comparison with other RL methods. 
We consider the $h$-value of an arbitrarily given single
$(s,a)$-pair at time $t$. It is determined by the sequence of rewards
$\lambda_{j+1}$ ($j=0,$ $\!\ldots,$ $\!t$ $\!-$ $\!1$) and times of visits,
which can be described using the sequence of Kronecker delta symbols $\delta_j$.
It is convenient to define the parameters
\begin{subequations}
\label{paras}
\begin{eqnarray}
  \bar{\eta}&:=&1-\eta,
  \\
 \bar{\gamma}&:=&1-\gamma,
  \\
  \chi&:=&\frac{\bar{\eta}}{\bar{\gamma}}.
\end{eqnarray}
\end{subequations}
With them, we can express the recursion (\ref{hur}) combined with (\ref{gur1})
in explicit form at time step $t$ (see App.~\ref{app:A1} for details),
\begin{equation}
\label{total}
  h_t=h_t^{\mathrm{res}}+h_t^{\mathrm{exp}}.
\end{equation}
The first term
\begin{equation}
\label{hres}
  h_t^{\mathrm{res}}=\bar{\gamma}^th_0+\bigl(1-\bar{\gamma}^t\bigr)
  h^{\mathrm{eq}}
\end{equation}
describes a transition from the initial $h_0$ to the asymptotic value
$h^{\mathrm{eq}}$. In particular, $h_t^{\mathrm{res}}$ $\!=$ $\!h_0$ holds
exactly for $\bar{\gamma}$ $\!=$ $\!1$ or for $h_0$ $\!=$ $\!h^{\mathrm{eq}}$,
and otherwise $h_t^{\mathrm{res}}$ $\!\approx$ $\!h^{\mathrm{eq}}$ holds
asymptotically for times long enough such that $\bar{\gamma}^t$ $\!\ll$ $\!1$.
This reward-independent term $h_t^{\mathrm{res}}$ is always present in
(\ref{total}), and (\ref{total}) reduces to it if the agent never receives any
reward, i.e., if the agent is at ``rest''. [Note that in case of an exponential policy function
(\ref{equ:softmax}), $h_t^{\mathrm{res}}$ has no effect on the policy, but this
is not of concern for the present discussion.] The second term in (\ref{total})
encodes the agent's ``experience'' and is determined by the history of visits
and rewards. Let us refer to time step $t$ $\!+$ $\!1$ for convenience,
\begin{subequations}
\label{expreps}
\begin{eqnarray}
\label{exprep1}
  h_{t+1}^{\mathrm{exp}}&=&\sum_{k=l_1}^{t}
  \left[\bar{\gamma}^{t-k}\bar{\eta}^{k-l(k)}\right]\lambda_{k+1}
  \\
\label{exprep2}
  &=&\sum_{j=1}^{j_t}
 \bar{\gamma}^{t-l_j}G\Bigl[l_j:\min(l_{j+1}-1,t),\chi\Bigr]
  \\
\label{exprep3}
  &=&\left(\sum_{j=1}^{j_t}\bar{\gamma}^{t-l_j}
  -\sum_{j=2}^{j_t}\chi^{l_j-l_{j-1}}\right)G(l_j:t,\chi).\hspace{0.5cm}
\end{eqnarray}
\end{subequations}
Here, $l_j$, $j$ $\!=$ $\!1,\ldots,j_t$, are the times at which the respective
edge is visited, i.e., $1$ $\!\le$ $\!l_1$ $\!\le$ $\!l_2$ $\!\le$ $\!\ldots$
$\!\le$ $\!l_{j_t}$ $\!\le$ $\!t$, and $j_t$ is thus the number of visits up to
time $t$. In (\ref{exprep1}), $l(k)$ is the time of the last visit with respect
to time step $k$, i.e., if $l_j$ $\!\le$ $\!k$ $\!<$ $\!l_{j+1}$, then
$l(k)$ $\!=$ $\!l_j$. Consequently, $k-l(k)$ is the number of steps that have
passed at time $k$ since the last visit occurred.
In (\ref{exprep2}) we have defined a truncated discounted return
\begin{equation}
\label{tdr}
  G\left(t:t+T,\,\chi\right):=\sum_{k=0}^{T}\chi^k\lambda_{t+k+1},
\end{equation}
which obeys
\begin{equation}
\label{tdr1}
  G(t_1:t_2,\,\chi)=\lambda_{t_1+1}+\chi\,{G}(t_1+1:t_2,\,\chi)
\end{equation}
and more generally
\begin{equation}
\label{tdr2}
  G(t_1:t_3,\,\chi)=G(t_1:t_2-1,\,\chi)
  +\chi^{t_2-t_1}G(t_2:t_3,\,\chi).
\end{equation}
Note that in (\ref{tdr}), discounting starts at the respective $t$ and not at
$t$ $\!=$ $\!0$, hence
\begin{equation}
\label{tdr3}
  \sum_{k=t_1}^{t_2}\chi^k\lambda_{k+1}=\chi^{t_1}G(t_1:t_2,\,\chi).
\end{equation}
\subsection{Accumulating glow}
In the following, we introduce accumulating glow, which is defined by the following update:
\begin{equation}
\label{gur2}
  g_{t+1}=(1-\eta)g_t+\delta_{t+1},
\end{equation}
 where each visit adds a value of 1 to
the current glow value of the respective edge. Writing the recursion (\ref{hur})
combined with (\ref{gur2}) instead of (\ref{gur1}) in explicit form yields
\begin{subequations}
\label{expacs}
\begin{eqnarray}
\label{expac1}
  h_{t+1}^{\mathrm{exp}}&=&\sum_{k=0}^{t}\Bigl[
  \bar{\gamma}^{t-k}\sum_{j=1}^{(l_j\le{k})}\bar{\eta}^{k-l_j}\Bigr]
  \lambda_{k+1}
  \\
\label{expac2}
  &=&\sum_{j=1}^{j_t}
  \bar{\gamma}^{t-l_j}G(l_j:t,\chi)
\end{eqnarray}
\end{subequations}
instead of (\ref{expreps}). The difference is that the subtracted sum in 
(\ref{exprep3}), which represents ``multiple re-visits'' is not included in
(\ref{expac2}).
\subsection{\label{app:A1}
Derivation of the explicit expressions for the $h$-value}
Writing the recursion (\ref{hur}) in explicit form gives
\begin{equation}
  h_n=h_n^{\mathrm{res}}+h_n^{\mathrm{exp}},
\end{equation}
which corresponds to (\ref{total}) with $h_n^{\mathrm{res}}$ given by
(\ref{hres}) and
\begin{equation}
\label{hurexp}
  h_n^{\mathrm{exp}}=\sum_{k=0}^{n-1}\bar{\gamma}^{n-k-1}g_k\lambda_{k+1}.
\end{equation}

\subsubsection{\label{rg}
Replacing glow}
The recursion (\ref{gur1}) for replacing glow yields the explicit expression
\begin{equation}
\label{gur1exp0}
  g_n=\bar{\eta}^n{g}_0\prod_{j=1}^n\bar{\delta}_j
  +\sum_{k=1}^{n}
  \bar{\eta}^{n-k}\delta_k\prod_{j=k+1}^n\bar{\delta}_j,
\end{equation}
where we have defined
\begin{equation}
  \bar{\delta}_j:=1-\delta_j
\end{equation}
for convenience. Setting $g_0$ $\!=$ $\!0$ gives
\begin{equation}
\label{gur1exp}
  g_n=\sum_{k=1}^{n}\bar{\eta}^{n-k}\Delta(k,n),
\end{equation}
where
\begin{equation}
  \Delta(k,n):=\delta_k\prod_{j=k+1}^n\bar{\delta}_j
\end{equation}
is 1 if the last visit occurred at step $k$ [$k$ $\!\le$ $\!n$, i.e., by
definition $\Delta(n,n) $:= $\!1$] and 0 otherwise. Together with
(\ref{hurexp}) we obtain after renaming $n$ as $t$ $\!+$ $\!1$
\begin{eqnarray}
\label{hurexp1}
  h_{t+1}^{\mathrm{exp}}&=&\sum_{k=0}^{t}\bar{\gamma}^{t-k}g_k\lambda_{k+1}
  \\
  &=&\sum_{k=0}^{t}\sum_{l=1}^{k}\bar{\gamma}^{t-k}\bar{\eta}^{k-l}
  \Delta(l,k)\lambda_{k+1}.
\end{eqnarray}
Rearranging summation
($\sum_{k=0}^{t}\sum_{l=1}^{k}$ $\!=$ $\!\sum_{l=1}^{t}\sum_{k=l}^{t}$),
applying (\ref{tdr}) together with (\ref{tdr2}) and (\ref{tdr3}), and resolving
the Kronecker delta symbols then gives (\ref{expreps}).
\subsubsection{Accumulating glow}
Similarly, the recursion (\ref{gur2}) for accumulating glow yields the explicit
expression
\begin{equation}
\label{gur2exp}
  g_n=\bar{\eta}^ng_0
  +\sum_{k=1}^{n}\bar{\eta}^{n-k}\delta_{k},
\end{equation}
where we set again $g_0$ $\!=$ $\!0$. Together with (\ref{hurexp}) we obtain
after renaming $n$ as $t$ $\!+$ $\!1$ the same expression as (\ref{hurexp1}),
from which (\ref{gur2exp2}) and (\ref{gur2exp1}) follow. In (\ref{gur2exp1}) we
have again rearranged summation and applied (\ref{tdr}) together with
(\ref{tdr3}). Resolving the Kronecker delta symbols then gives (\ref{expacs}).
\subsection{\label{app:A1.3}
Order in glow updating}
Note that in the updating of the replacing edge glow applied in
\cite{Mau15}, the glow value of visited edges is first reset to one,
followed by a damping of all edges by multiplying $g$ with $\bar{\eta}$.
In contrast, the recursion (\ref{gur1}) for replacing glow applies the damping
first and then resets the glow value of visited edges to 1. We may understand
(\ref{gur1}) as first making up for the damping of the previous step and then do
the actual resetting of the present step. As an embedding description we may
define an $s$-ordered replacing glow update
\begin{equation}
\label{gur1o}
 g_{t+1}=s\delta_{t+1}+\bar{\eta}\bar{\delta}_{t+1}g_t
\end{equation}
generalizing (\ref{gur1}), where $s$ is a real valued ordering parameter.
$s$ $\!=$ $\!\bar{\eta}$ describes the case of 1. resetting and 2. damping as
done in \cite{Mau15,Melnikov2018}, whereas $s$ $\!=$ $\!1$ describes the case of
1. damping and 2. resetting as done in (\ref{gur1}). In explicit form,
(\ref{gur1o}) yields
\begin{equation}
\label{gur1oex}
  g_n=\bar{\eta}^n{g}_0\prod_{j=1}^n\bar{\delta}_j
  +s\sum_{k=1}^{n}
  \bar{\eta}^{n-k}\delta_k\prod_{j=k+1}^n\bar{\delta}_j
\end{equation}
instead of (\ref{gur1exp0}). Analogously, an $s$-ordered accumulating glow
update
\begin{equation}
\label{gur2o}
 g_{t+1}=s\delta_{t+1}+\bar{\eta}g_t
\end{equation}
generalizes (\ref{gur2}) by describing 1. incrementing and 2. damping for
$s$ $\!=$ $\!\bar{\eta}$ and 1. damping and 2. incrementing for $s$ $\!=$ $\!1$
as done in (\ref{gur2}). The explicit form of (\ref{gur2o}) is
\begin{equation}
\label{gur2oex}
  g_n=\bar{\eta}^ng_0
  +s\sum_{k=1}^{n}\bar{\eta}^{n-k}\delta_{k}
\end{equation}
instead of (\ref{gur2exp}). The only difference is the extra factor $s$ in the
second term in (\ref{gur1oex}) and (\ref{gur2oex}) compared to
(\ref{gur1exp0}) and (\ref{gur2exp}), respectively, with which the
$h_{t+1}^{\mathrm{exp}}$ in (\ref{hurexp1}) (which holds for both types of glow)
and hence in (\ref{expreps}) and (\ref{expacs}) would have to be
multiplied. The difference is therefore minor and irrelevant for our
considerations.

\section{Comparative Analysis of Projective Simulation and other RL methods}
\label{app:comp}

A specific contribution, which the PS-updates have to offer to RL consists
in supplementing the usual forward discounting with a backward discounting
enabled by the damping of the $(s,a)$-pair values, which amounts to a
generalization of the standard notion of return. On the other hand, the
incompatibility of discounting with function approximation mentioned in
App.~\ref{app:RLmethods} may also extend to damping.

In the following discussion, we want to analyze the difference between PS and other RL methods. 
The first observation is that neither (\ref{expreps}) nor (\ref{expacs})
update averages, instead they add ``double-discounted'' rewards.
In what follows, first we show in App.~\ref{sec:eta} how averaging can be implemented before we show in App.~\ref{sec:ddisc} some simple effects of
forward and backward discounting, assuming that averaging is carried out.
Averaging will not be integrated into the PS, as we do not want to give up the simplicity of the PS updates. 
This discussion merely serves as a thorough analysis of the differences between PS and methods that use averaging. 

In the language of App.~\ref{app:RLmethods}, the basic PS updates (\ref{hur}) constitute a
tabular model-free on-policy online learning method. In the analysis in App.~\ref{sec:compare}, we show that among the methods in
App.~\ref{app:RLmethods}, it is tabular SARSA($\lambda$) defined in
(\ref{uq1})--(\ref{uq2}), which comes closest to (\ref{hur}), because it has an
eligibility value $z(s,a)$ ascribed to each $(s,a)$-pair that is the counterpart
of the respective glow value $g(s,a)$ and a trace decay parameter $\lambda$,
which may be ``meta-learned'' (i.e., optimized). 
Thus, in App.\ref{sec:compare} we analyze the differences and similarities between PS and SARSA.

\subsubsection{\label{sec:eta}
Implementing a temporal averaging}
In this section, we show how temporal averaging can be integrated by adding to
the $h$- and $g$-value a third variable $N$ $\!=$ $\!N(s,a)$ to each
$(s,a)$-pair, which counts the number of visits by updating it according to
\begin{equation}
\label{Nur}
  N_{t+1}=N_t+\delta_{t+1},
\end{equation}
which is formally the same update as (\ref{gur2}) for undamped accumulating
glow. With it, we could consider the normalized
$\tilde{h}_{t}$ $\!=$ $\!{h}_{t}/N_{t}$ and initialize with $N_{0}$ $\!=$ $\!1$
to avoid division by zero, so that explicitly $N_{t}$ $\!=$ $\!N_0$ $\!+$
$\!\sum_{k=1}^{t}\delta_k$ $\!=$ $\!j_t$ $\!+$ $\!1$. Alternatively, we can
integrate the normalization into the update rule $f$,
${h}_{t+1}$ $\!=$ $\!N_{t+1}\tilde{h}_{t+1}$ $\!=$ $\!f({h}_{t})$ $\!=$
$\!f(N_{t}\tilde{h}_{t})$ by replacing (\ref{hur}) with
\begin{eqnarray}
  \tilde{h}_{t+1}&=&\frac{\alpha_{t+1}}{\alpha_{t}}\tilde{h}_t
  -\gamma\left(\frac{\alpha_{t+1}}{\alpha_{t}}\tilde{h}_t
  -\alpha_{t+1}{h}^{\mathrm{eq}}\right)
  +\alpha_{t+1}g_t\lambda_{t+1}
  \nonumber\\
\label{hurtilde}
  &\approx&\tilde{h}_t-\gamma(\tilde{h}_t
  -\alpha_{t}{h}^{\mathrm{eq}})
  +\alpha_{t}g_t\lambda_{t+1},
  \\
\label{aur}
 \alpha_{t+1}&=&\frac{\alpha_{t}}{1+\alpha_{t}\delta_{t+1}},
\end{eqnarray}
where the approximation (\ref{hurtilde}) holds as soon as
$\alpha_{t}$ $\!\ll$ $\!1$. Instead of $N$, we thus then keep for each
$(s,a)$-pair a separate time-dependent learning rate
$\alpha_t$ $\!=$ $\!N_t^{-1}$ $\!=$ $\!\alpha_t(s,a)$ and update it according to
(\ref{aur}). 

For accumulating glow, (\ref{expac2}) sums over all visits $j$ the
backward-discounted returns that follow these visits up to the present $t$, and
$\tilde{h}_{t+1}^{\mathrm{exp}}$ thus becomes (for large $t$) an estimate of the
average backward-discounted return that follows a visit. In contrast, the
first-visit counterpart (\ref{PS1prima}) only depends on the time $l_1$ of the first
visit, which is analytically more easily analyzed in an episodic environment, where after each episode, the glow values of all $(s,a)$-pairs are reset to zero.

The updates involving (\ref{Nur}) or (\ref{aur}) may be read as a laborious
reinvention of an online approximation of an every-visit Monte Carlo method, but
provide the following insight: For the action value methods in the context of
Sec.~\ref{app:RLmethods}, the learning rate can in practice (especially when dealing with
deterministic environments) often be kept constant rather than gradually
decreasing it, where the precise value of this constant doesn't matter. For our
updates of $\tilde{h}$, omitting the correction by $N$ or $\alpha$ and working
with the original $h$ should work reasonably well, too, in such problems.

\subsubsection{Effect of double discounting on a temporal average}
\label{sec:ddisc}
As an elementary illustration of the effect of forward discounting via
$\bar{\eta}$ and backward discounting via $\bar{\gamma}$ on agent learning
consider a weighted arithmetic mean
\begin{equation}
\label{wam}
  \bar{x}^{(t)}=\frac{\sum_{k=1}^{t}w_kx_k}{\sum_{k=1}^{t}w_k}
\end{equation}
of random samples $x_k$ with variable but known weights $w_k$ $\!\ge$ $\!0$
($w_1$ $\!>$ $\!0$). If the samples are drawn in succession
$x_1,$ $\!x_2,$ $\!\ldots$, then the average can be updated incrementally,
\begin{equation}
  \bar{x}^{(t)}=(1-\alpha_t)\bar{x}^{(t-1)}+\alpha_tx_t,
\end{equation}
with a ``learning rate''
\begin{equation}
  \alpha_t=\frac{w_t}{\sum_{k=1}^{t}w_k},
\end{equation}
which in general fluctuates within $\alpha_t$ $\!\in$ $\![0,1)$ depending on the
weight sequence $w_1,$ $\!w_2,$ $\!\ldots$. (Note that an incremental
formulation
\begin{equation}
  \alpha_{t+1}=\left(1+\frac{w_t}{w_{t+1}\alpha_t}\right)^{-1}
\end{equation}
would require that $w_k$ $\!>$ $\!0$ holds $\forall$ $\!k$.) Of particular
interest for our discussion is an exponential choice of weights
\begin{equation}
\label{eam}
  w_k=w^k,\quad\alpha_t=\frac{1-w^{-1}}{1-w^{-t}}.
\end{equation}
In (\ref{eam}) we can distinguish the following cases:

(a) For $w$ $\!=$ $\!1$ all samples are given equal weight and the learning rate
$\alpha_t$ $\!=$ $\!t^{-1}$ $\!\stackrel{t\to\infty}{\to}$ $\!0$ decays to zero
in a manner of (\ref{cc}). In the special case, when the $x_k$ are drawn from
i.i.d. random variables $X_k$ $\!\equiv$ $\!X$ with variance
$\sigma^2(X)$ $\!=$ $\!\sigma^2$, the total variance
$\sigma^2_{(t)}$ $\!=$ $\!t^{-1}\sigma^2$
of $\frac{\sum_{k=1}^{t}w_kX_k}{\sum_{k=1}^{t}w_k}$
vanishes with growing $t$ and $\bar{x}^{(t)}$ converges to the expectation value
$E(X)$. In the context of agent learning, we may interpret $\bar{x}^{(t)}$ as
the agent's experience (e.g., a current value estimate of some given
state-action-pair). If after some longer time $t$ $\!\gg$ $\!1$, the environment
statistics changes ($X_k$ $\!\equiv$ $\!X$  no longer holds), the average
$\bar{x}^{(t)}$ will start to change only slowly. 

(b) The case $w$ $\!<$ $\!1$ in (\ref{eam}) corresponds to the effect of a
discounting from the beginning of learning towards the present $t$ by the factor
$\bar{\eta}^{k}$. The learning rate
$\alpha_t$ $\!\stackrel{t\gg1}{\approx}$ $\!(w^{-1}-1)w^{t}$
$\!\stackrel{t\to\infty}{\to}$ $\!0$ decays to zero exponentially and the agent
will cease to learn anything after some finite time period of the order
$\Delta{t}$ $\!\approx$ $\!-(\ln{w})^{-1}$ due to decay of the weights in
(\ref{eam}). After that time, the agent will behave solely according to this
early experience ``imprinted'' into it.

(c) The case $w$ $\!>$ $\!1$ in (\ref{eam}) corresponds to the effect of
discounting from the present $t$ towards the past by a damping factor
$\bar{\gamma}^{t-k}$. The learning rate
$\alpha_t$ $\!\stackrel{t\to\infty}{\to}$ $\!1-w^{-1}$ converges to a
positive constant and the agent remains ``fluid'' in its ability to react to
changes in the environment statistics. On the other hand, since it's remembered
experience reaches only a time period of the order
$\Delta{t}$ $\!\approx$ $\!(\ln{w})^{-1}$ from the present into the past, such
an agent will just ``chase the latest trend'' without learning anything
properly.

In the special case, when the $x_k$ are drawn from i.i.d. random variables
$X_k$ $\!\equiv$ $\!X$ with variance $\sigma^2(X)$ $\!=$ $\!\sigma^2$, the total
variance 
$\sigma^2_{(t)}$ $\!\stackrel{t\gg1}{\approx}$ $\!\frac{|w-1|}{w+1}\sigma^2$
of $\frac{\sum_{k=1}^{t}w_kX_k}{\sum_{k=1}^{t}w_k}$
converges to a positive constant in both cases (b) and (c), that is, when
$w$ $\!\neq$ $\!1$. The difference is that in case (c), the weighted mean
$\bar{x}^{(t)}$ keeps fluctuating, whereas in case (b), this variance has been
``crystallized'' into a bias $\bar{x}^{(t)}$ $\!-$ $\!E(X)$ of the 
early experience $\bar{x}^{(t)}$ which is fixed by the samples in (\ref{wam})
with respect to the actual $E(X)$.


\subsubsection{Description of a formal ensemble average}
We restrict attention to the simpler accumulating glow (\ref{expacs}), which we
rewrite with the Kronecker delta symbols kept explicitly,
\begin{eqnarray}
\label{gur2exp2}
  h_{t+1}^{\mathrm{exp}}
  &=&\sum_{k=0}^{t}\sum_{l=1}^{k}\bar{\gamma}^{t-k}\bar{\eta}^{k-l}
  \lambda_{k+1}\delta_l
  \\
\label{gur2exp1}
  &=&\sum_{l=1}^{t}\bar{\gamma}^{t-l}G(l:t,\chi)\delta_l
\end{eqnarray}
(see App.~\ref{rg} for the corresponding expressions describing replacing glow).
Each $\delta_l$ samples the ``occupation'' of the given $(s,a)$-pair at time
$l$, whose probability is given by $p_l(s,a)$. For an ensemble of independent
agents running in parallel, we can thus replace the $\delta_l$ with these
probabilities and write
\begin{eqnarray}
\label{gur2exp3}
  \left\langle{h}_{t+1}^{\mathrm{exp}}\right\rangle_{\mathrm{ens}}
  &=&\sum_{l=1}^{t}\bar{\gamma}^{t-l}G(l:t,\chi)p_l
  \\
\label{gur2exp4}
  &=&\left\langle\bar{\gamma}^{t-l}G(l:t,\chi)\right\rangle_{l=1,\ldots,t}.
\end{eqnarray}
While (\ref{gur2exp1}) sums for all times $l$ the respective backward-discounted
return $\bar{\gamma}^{t-l}G(l:t,\chi)$ from that time \emph{under the condition
that} the edge was visited, the ensemble average (\ref{gur2exp3}) performs an
average with respect to the $p_l$ over all times $l$ up to the present $t$. The
problem is that we do not know the distribution $p_l$, which itself is
affected by both the environment and the agent's policy.

What we can do, however, is to consider the average return \emph{that follows a
visit} at given time $l$. The average number $n_l$ $\!=$ $\!Np_l$ of visits per
unit of time at time $l$ is for an ensemble of size $N$ just given by $p_l$,
with which we normalize each summand in (\ref{gur2exp3}). The sum over all times
$l$ of these average returns per visit with respect to time $l$ can then be
written as
\begin{eqnarray}
\label{gur2exp5}
  \widetilde{\left\langle{h}_{t+1}^{\mathrm{exp}}\right\rangle}_{\mathrm{ens}}
  &=&\frac{1}{N}\sum_{l=1}^{t}\frac{\bar{\gamma}^{t-l}G(l:t,\chi)p_l}{p_l}
  \\
\label{gur2exp6}
  &=&\frac{1}{N}\sum_{l=1}^{t}\bar{\gamma}^{t-l}G(l:t,\chi)
  \\
\label{gur2exp7}
  &=&\frac{1}{N}\frac{\bar{\gamma}^{t}}{1-\bar{\eta}}
  \left[G(0:t,\bar{\gamma}^{-1})-G(0:t,\chi)\right].\quad
\end{eqnarray}
The normalization in (\ref{gur2exp5}) is analogous to the one that motivated
the logarithm in (\ref{ac-e}) as discussed in App.~\ref{app:RLmethods} and
\ref{sec:ddisc}: it makes the expression independent of the state
distribution $p_l$. It also reveals that what we have called
``double discounting'', i.e., the convolution (\ref{gur2exp6})
of the return $G(l:t,\chi)$ with the exponential $\bar{\gamma}^{l}$ amounts to
the difference (\ref{gur2exp7}) between two returns from the beginning
$t$ $\!=$ $\!0$.

For a single agent in Sec.~\ref{sec:eta}, there cannot be more than one visit
at each time $l$. We therefore had to take the sum ${h}_{t+1}^{\mathrm{exp}}$ at
time $t$, and divide it by the total (cumulative) number of visits $N_{t}$ that
occurred up to this time, to get an estimate
$\tilde{h}_{t}$ $\!=$ $\!{h}_{t}/N_{t}$ of the average return per visit. One
possibility to implement (\ref{gur2exp5}) for a single agent consists in
training the agent ``off-policy'' by separating exploration and exploitation,
which can be done by choosing the softmax policy (\ref{equ:softmax}). During
periods of exploration (e.g. if the agent is not needed), we choose
a small $\beta$ , whereas for exploitation, we temporarily disable the updating
(\ref{hur}) and switch to a large $\beta$. By large (small) we mean values of
$\beta$ such that for all $x$ $\!=$ $\!h_{ij}$ encountered in
(\ref{equ:softmax}), the argument of the exponential obeys
$\beta{x}$ $\!\gg$ $\!1$ ($\beta{x}$ $\!\ll$ $\!1$). For graphs that have for
symmetry reasons the property that $p_l$ $\!\equiv$ $\!p$ for a random walker
(note that for ergodic MDPs the $p_l$ eventually become independent of the
initial conditions), we should be able to realize (\ref{gur2exp5}) during the
periods of exploration. It is clear, that this is impractical for all but small
finite MDPs.

\subsubsection{Relation of the PS-updates to other RL methods}
\label{sec:compare}
In this section, we compare PS to the standard RL methods presented in App.~\ref{app:RLmethods}.
One may interpret the PS-updates (\ref{hur}) as implementing a direct policy
method. On the other hand, these updates do not involve gradients. 
To draw connections between PS and direct policy methods, we consider the gradient
$\bm{\nabla}p$ of the probability $p_{ij}$ $\!=$ $\!p(a_j|s_i)$ of selecting
action $a_j$ in state $s_i$, i.e., one element of the policy $\pi$ and restrict
to our case (\ref{equ:policy}), i.e.,
$p_{ij}$ $\!=$ $\!\frac{\Pi({h}_{ij})}{\kappa_i}$. As in the derivation
(\ref{Kmatrix}) of tabular SARSA, we identify the components of the parameter
vector $\bm{\theta}$ (with respect to which we want to determine the gradient)
with the edges ${h}_{kl}$. The gradient of $p$ thus becomes a matrix, whose
element $kl$ reads
\begin{equation}
\label{dp}
  (\bm{\nabla}p_{ij})_{kl}=\frac{\partial{p}_{ij}}{\partial{h}_{kl}}
  =\frac{\Pi^\prime({h}_{il})}{\kappa_i}
  \left[\delta_{ki}\delta_{lj}
  -\frac{\Pi({h}_{ij})}{\kappa_i}\delta_{ki}\right],
\end{equation}
where $\Pi^\prime(x)$ $\!=$ $\!\frac{{d}\Pi(x)}{{d}x}$. To obtain
$\bm{\nabla}\ln{p}$, we just multiply this with the factor
$p_{ij}^{-1}$ $\!=$ $\!\frac{\kappa_i}{\Pi({h}_{ij})}$. The term 
$\delta_{ki}\delta_{lj}$ is also present in the PS-update, where it corresponds
to the strengthening of a visited edge. The subtracted second term proportional
to $\delta_{ki}$ represents a weakening of all edges connecting $s_i$, which is not present in the PS-update. 

With (\ref{dp}) we can now compare the PS-updates (\ref{hur}) with the methods
in Fig.~\ref{fig3}. Among the action value methods, it is tabular
SARSA($\lambda$) defined in (\ref{uq1})--(\ref{uq2}), which resembles the
PS-updates most. Let us rewrite the SARSA-updates in the notation used within
this. After renaming the reward $R$ as $\lambda$, the action value
function $q$ as $h$, the eligibility vector $\bm{z}$ as (matrix) $g$, the
discount rate $\gamma$ as $\gamma_{\mathrm{dis}}$, and the trace-decay rate
$\lambda$ as $\lambda_{\mathrm{tra}}$ for clarity, the TD-error (\ref{TDe})
reads
\begin{equation}
\label{TDePS}
  \delta^{\mathrm{TD}}_t
  =\lambda_{t+1}+\gamma_{\mathrm{dis}}
  {h}_t(s_{t+1},a_{t+1})-{h}_t(s_t,a_t),
\end{equation}
with which (\ref{uq1})--(\ref{uq2}) for SARSA($\lambda_{\mathrm{tra}}$) become
\begin{eqnarray}
\label{uq1PS}
  {h}_{t+1}&=&{h}_{t}+\alpha\delta^{\mathrm{TD}}_t{g}_{t},
  \\
\label{uq2PS}
  {g}_{t+1}&=&\gamma_{\mathrm{dis}}\lambda_{\mathrm{tra}}{g}_{t}
  +\bm{\delta}_{t+1},
\end{eqnarray}
where $\bm{\delta}_t$ is a matrix of Kronecker deltas describing which of the
$(s,a)$-pairs has been visited at time $t$. A tabular direct policy method
follows in the same way from (\ref{ac-a}), (\ref{ac-e}), and (\ref{ac-f})
(setting again $\Gamma$ $\!=$ $\!1$): (\ref{TDePS}) and (\ref{uq1PS}) remain
identical, merely (\ref{uq2PS}) is replaced with
\begin{equation}
\label{gur2s}
  {g}_{t+1}=\gamma_{\mathrm{dis}}\lambda_{\mathrm{tra}}{g}_{t}
  +(\bm{\nabla}\ln{p})_{t+1},
\end{equation}
where for $\bm{\nabla}\ln{p}$ we substitute (\ref{dp}) together with the extra
factor as explained in the fext following (\ref{dp}). While (\ref{uq2PS})
recovers the update (\ref{gur2}) for accumulating glow [(\ref{gur2}) considers a
given $(s,a)$-pair, (\ref{uq2PS}) the whole matrix], (\ref{gur2s}) is in fact
even more complex than (\ref{uq2PS}). 

One important difference is the presence of the term ${h}_t(s_{t+1},a_{t+1})$ in
(\ref{TDePS}) which persists even if we disable bootstrapping by setting
$\lambda_{\mathrm{tra}}$ $\!=$ $\!1$. We can also rewrite SARSA in the
``local'' fashion of the PS-updates (\ref{hur}), which we here repeat as
\begin{equation}
\label{hur2}
  h_{t+1}=h_t+\lambda_{t+1}g_t-\gamma{h}_t+\gamma{h}^{\mathrm{eq}}
\end{equation}
for convenience. To rewrite SARSA($\lambda_{\mathrm{tra}}$) in the form of
(\ref{hur2}), we proceed as in the justification of accumulating traces outlined
in App.~\ref{app:A-1}. First, we sum all increments in (\ref{uq1PS}) up to some
time $T$, i.e., ${h}_{T}$ $\!=$ $\!{h}_{0}$ $\!+$
$\!\alpha\sum_{t=0}^{T-1}\delta^{\mathrm{TD}}_t{g}_{t}$, then rewrite
the term involving ${h}_t(s_{t+1},a_{t+1})$ in $\delta^{\mathrm{TD}}_t$ as
$\sum_{t=0}^{T-1}{h}_t(s_{t+1},a_{t+1}){g}_{t}$ $\!=$
$\!\sum_{t=1}^{T}{h}_{t-1}(s_{t},a_{t}){g}_{t-1}$ and substitute
${g}_{t-1}$ $\!=$
$\!\frac{{g}_{t}-\delta_t}{\gamma_{\mathrm{dis}}\lambda_{\mathrm{tra}}}$.
If we now ignore (a) the change of the $h$-values over a single time step (which
holds for small $\alpha$),
${h}_{t-1}(s_{t},a_{t})$ $\!\approx$ $\!{h}_{t}(s_{t},a_{t})$,
and (b) ignore the shift of argument in the summation (i.e. ignore the first and
last sum term), then identifying each term referring to a given $t$ in the sum
over all increments with an individual update leads to a ``PS-style''
form of SARSA($\lambda_{\mathrm{tra}}$),
\begin{equation}
\label{sarsalocal}
  h_{t+1}=h_t+\alpha\lambda_{t+1}g_t
  -\alpha{h}_{t}(s_{t},a_{t})\bigl[
  \lambda_{\mathrm{tra}}^{-1}\bm{\delta}_{t}
  +(1-\lambda_{\mathrm{tra}}^{-1}){g}_{t}
  \bigr],
\end{equation}
in which $\gamma_{\mathrm{dis}}$ no longer appears [it remains in
(\ref{uq2PS})]. 

We can simplify Equation (\ref{sarsalocal}) if we disable bootstrapping by setting
$\lambda_{\mathrm{tra}}$ $\!=$ $\!1$, so that it becomes even more similar to
PS. On the other hand, if we take into account that PS does not use averaging, PS carries some similarities 
to (an online approximation of) Monte Carlo approaches. 
The type of glow then determines the corresponding type of Monte
Carlo method. For example, using replacing glow relates it more to first-visit
Monte Carlo, whereas accumulating glow relates it more to every-visit Monte Carlo.

\section{\label{app:details} Mathematical details of the convergence proof}

In this appendix, we provide the mathematical details we skipped during the proof of Theorem \ref{lemma} in Sec.~\ref{sect:convergence}. We are left with showing that $\alpha_m(\edge)$, given in Eq.~(\ref{equ.alpha_m}) satisfies Condition 2 in Theorem \ref{theorem}, and $F_m(\edge)$, given in Eq.~(\ref{eq:F_definition}), satisfies Conditions 3 and 4. From these, Condition 3 is the most involving, and to some extent is the core of the proof. Condition 4 follows trivially under our assumption of bounded rewards. One can easily see that bounded rewards imply that $\htilde$ values are upper and lower bounded. Given that optimal $Q$-values are bounded as well it follows that $\mathrm{Var}\{F_m(\edge)|P_m\}\leq K'$ for some constant $K'$. The remaining two properties are proven in the following.

\subsection{Proving that $\alpha_m(e)$ satisfies Condition 2 in Theorem \ref{theorem}}

Let us recall that 
\begin{equation}\label{equ.alpha_app}
\alpha_m (\edge)\coloneqq\frac{\chi_m(\edge)}{N_m(\edge)+1}=\frac{\chi_m (\edge)}{\sum_{j=1}^{m} \chi_j (\edge) +1},
\end{equation}
where the $\chi_m(\edge)$ are given by 
\begin{equation}
\chi_m(\edge)=
\begin{cases}
1\quad\textrm{if $\edge$ was visited during the $m$th episode,}\\
0\quad\textrm{otherwise.}
\end{cases}
\end{equation}
Due to the fact that the policy guarantees infinite exploration, we know that the number of non-zero terms from the sequence $\mathcal{Q}_1\coloneqq\{\alpha_m(\edge)\}_{1\leq m < \infty}$ is infinite. Thus let $\mathcal{Q}_2\coloneqq \{\tilde{\alpha}_n (\edge)\}_{1\leq n < \infty}$ be the subsequence of $\mathcal{Q}_1$ obtained by removing all zero elements, and relabeling the remaining ones as $1,2,3$, etc. Clearly, we have that 
\begin{eqnarray}
\sum_{m=1}^{\infty} \alpha_m (\edge)  = \sum_{n=1}^{\infty} \tilde{\alpha}_n (\edge),\label{equ.equality1_app}\\
\sum_{m=1}^{\infty} [\alpha_m(\edge)]^2  = \sum_{n=1}^{\infty} [\tilde{\alpha}_n (\edge)]^2\label{equ.equality2_app}.
\end{eqnarray}
Furthermore, it is trivial to see that the non-zero terms $\tilde{\alpha}_n(\edge)=1/(n+1)$. Given that $\sum_n 1/n=\infty$ and $\sum_n 1/n^2 <\infty$, it follows that 
\begin{eqnarray}
\label{cc1}
\sum_{m=1}^{\infty}\alpha_{m}(\edge) & =  \infty, \label{equ.alpha_1_app}\\
\label{cc2}
\sum_{m=1}^{\infty}\alpha^2_{m}(\edge)  &<  \infty, \label{equ.alpha_2_app}
\end{eqnarray}
as we wanted to prove.
\subsection{Contraction of $F_{m}(e)$}
In this part of the appendix we show that in the case where the glow parameter of the PS model $\etabar$ is set equal to the discount parameter $\gammadis$ associated to the MDP, $F_m(\edge)\coloneqq D_m(\edge)-\qstar(\edge) $ satisfies
\begin{equation}\label{equ.to_prove}
	\normw{\E\{F_m(\cdot)|P_m\}}\leq f(\gammadis) \normw{\Delta_m (\cdot)}+c_m,
\end{equation}
where $\Delta_{m}(e)\coloneqq\tilde{h}_{m}(e)-\qstar(e)$, $f(\gammadis)=\frac{2\gammadis}{1-\gammadis}$ and $c_m$ converges to zero w.p.1.

In the update rule for $\Delta_m(\edge)$ given in Eq.~(\ref{equ.deltaupdate}), $F_m(\edge)$ appears multiplied by the learning rate coefficient $\alpha_m(\edge)$. Given that $\alpha_m(\edge)=0$ in the case where $\chi_m(\edge)=0$, we can, w.l.o.g., define $F_m(\edge)=0$ for that case. This is made explicit by the factor $\chi_m (\edge)$ in the definition of $F_m(\edge)$ given in Eq.~(\ref{eq:F_definition}), which for $\etabar=\gamma$ leads to
\begin{equation}\label{equ.F_m_appendix}
	F_m(\edge)=\chi_m (\edge) \left(\sum_{j=0}^{T_m-t_m(\edge)}\gamma^j \lambda_{t_m(\edge)+j} -\qstar(\edge)\right).
\end{equation}
Following this definition of $F_m(\edge)$ we have that 
\begin{equation}
	\E\{F_m(\edge)|P_m\}=\E\{F_m(\edge)|\chi_m(\edge)=1, P_m\}.
\end{equation} 
To simplify the notation, in the following we will always assume that $\edge$ has been visited at least once during episode $m$, but for simplicity in our notation we omit writing the condition on $\chi_m(\edge)=1$ in the expected value.

The past of the process at episode $m$,  $P_m$, includes every state, action, and reward received by the agent from $t=0$ until the beginning of the episode $m$. In particular it determines the set of $\htilde_m$ values, which in turn determine the policy at the beginning of the $m$th episode. For the clarity of this proof we will first consider the case where the policy is kept unchanged during episodes and only updated at the beginning of a new episode, shown that Eq.~(\ref{equ.to_prove}) holds under those assumptions. Later on we relax that condition and show that the differences accumulated by the policy during the episode converge to zero with probability one. This allows us to prove Eq.~(\ref{equ.to_prove}) also in the case where the policy is updated every time step.

\subsubsection{Constant policies during the episodes}

Given that the number of time steps required by an agent to hit a terminal state is unbounded, the number of terms in $F_m(\edge)$ could be arbitrarily large. Therefore, we construct a family of truncated versions of $F_m(\edge)$, where the maximum number of terms is upper bounded. Let us define
\begin{equation}\label{equ.F_m_truncated}
	\begin{aligned}
		F_m^{(k)}(\edge)\coloneqq&\sum_{j=0}^{k}\Theta(T_m-t_m(\edge)-j)\gammadis^j\lambda_{t_m(\edge)+j}\\
		+\Theta&(T_m-t_m(\edge)-k)\gammadis^{t+1}\htilde_m(S_{k+1},A_{k+1})-\qstar(\edge),
	\end{aligned}
\end{equation}
where $\Theta(l)=1$ for $l\geq 0$ and it is zero otherwise, and we have defined $\htilde(s_T,a)=0$, for all terminal states $s_T$\footnote{Since episodes end when a terminal state is reached, the PS agent does not need to store $h$-values associated with terminal states, thus we can define them equal to zero.}. Comparing Eqs.~(\ref{equ.F_m_appendix}) and (\ref{equ.F_m_truncated}) one can see that $F_m^{(k)}(\edge)=F_m(\edge)$ in the case where the agent takes less than $k$ time steps to finish the episode since $\edge$ is visited for the first time during the $m$th episode, i.e. whenever $T_m-t_m(\edge)<k$. Considering that the policy guarantees infinite exploration, the probability of not reaching a terminal state after $k$ time steps (during a single episode) decays exponentially with $k$, and therefore
\begin{equation}\label{equ.F_limit}
	\E\{F_m(\edge)|P_m\}=\lim_{k\rightarrow \infty} \E\{F_m^{(k)}(\edge)|P_m\}.
\end{equation}

In the following we construct an upper bound for $\E\{F_m^{(k)}(\edge)|P_m\}$ that holds for all $k$, and hence due to Eq.~(\ref{equ.F_limit}) it also bounds $\E\{F_m(\edge)|P_m\}$. We can write $F_m^{(k)}(\edge)$ in the following form 

\begin{widetext}
\begin{equation}\label{equ.F_truncated_expected}
	\begin{aligned}
		\E\{F_m^{(k)}(\edge)|P_m\}=&r(\edge)+\sum_{l=1}^{k} \sum_{\edge^{(1)},\dots \edge^{(l)}}  \mathrm{Pr}\left[\edge,\edge^{(1)},\dots, \edge^{(l)}|P_m\right]\,\gammadis ^lr(\edge^{(l)})\\
		&+\sum_{\edge^{(1)},\dots,\edge^{(k+1)}}  \mathrm{Pr}\left[\edge,\edge^{(1)},\dots, \edge^{(k+1)}|P_m\right]\,\gammadis ^{k+1} \htilde_m(\edge^{(k+1)})
		-\qstar(\edge),
	\end{aligned}
\end{equation}
\end{widetext}
where we used the short-hand notation $\mathrm{Pr} \left[\edge,\edge^{(1)},\dots,\edge^{(l)}|P_m\right]$ to denote the probability of an agent following the sequence of state-action pairs $\edge,\edge^{(1)},\dots,\edge^{(l)}$. Given that, for the moment, we are considering constant policies during episodes, these probabilities only depend on the episode index $m$. In addition, in order to have a simpler expression, w.l.o.g. we assume that transitions from a terminal state return with probability one to a terminal state with zero reward. Note that this assumption together with the fact that $\htilde_m(s_T,a)=0$ for all terminal states $s_T$, allows us to write the summations in Eq.~(\ref{equ.F_truncated_expected}) over all edges, including those associated to terminal states.

The following step consists in writing $\E\{F_m^{(k)}(\edge)|P_m\}$ as a recursive relation in $k$. Given that the policies are kept constant during episodes they satisfy that
\begin{equation}
	\begin{aligned}
		&\mathrm{Pr}\left[\edge,\edge^{(1)},\dots \edge^{(k+1)}|P_m\right]=\\
		&\quad\mathrm{Pr} \left[\edge,\edge^{(1)},\dots,\edge^{(k)}|P_m\right]\mathrm{Pr}\left[\edge^{(k)},\edge^{(k+1)}|P_m\right].
	\end{aligned}
\end{equation}
Plugging this equation into Eq.~(\ref{equ.F_truncated_expected}) and adding canceling terms we end up with the expression
\begin{widetext}
	\begin{equation}\label{equ.F_truncated_expected_2}
		\begin{aligned}
		\E\{F_m^{(k)}(\edge)|P_m\}=&r(\edge)+  \sum_{l=1}^{k-1} \sum_{\edge^{(1)},\dots, \edge^{(l)}}  \mathrm{Pr}\left[\edge,\edge^{(1)},\dots, \edge^{(l)}|P_m\right]\, \gammadis^l r(\edge^{(l)})\\
		&+ \sum_{\edge^{(1)},\dots,\edge^{(k)}}\mathrm{Pr}\left[\edge,\edge^{(1)},\dots, \edge^{(k)}|P_m\right]\,\gammadis ^k \htilde_m(\edge^{(k)})
		-\qstar(\edge)\\
		&+\sum_{\edge^{(1)},\dots, \edge^{(k)} }\mathrm{Pr} \left[\edge,\edge^{(1)},\dots,\edge^{(k)}|P_m\right]  \gammadis ^k \Big\{  -\htilde_m(\edge^{(k)})+\qstar(\edge^{(k)})\\
		& \phantom{\sum}+ r(\edge^{(k)})+\sum_{\edge^{(k+1)}}\mathrm{Pr} [\edge^{(k)},\edge^{(k+1)}|P_m]\, \gammadis \htilde_m(\edge^{(k+1)})-\qstar(\edge^{(k)}) \Big\}.
		\end{aligned}
	\end{equation}
\end{widetext}
Comparing Eqs.~(\ref{equ.F_truncated_expected}) and (\ref{equ.F_truncated_expected_2}) one can see that the first two lines in the previous equation equal $\E\{F_m^{(k-1)}(\edge)|P_m\}$. Furthermore, in the third and fourth lines, the quantities within curly brackets correspond to the definition of $\Delta_m(\edge^{(k)})$ and $\E\{F_m^{(0)}(\edge^{(k)})|P_m\}$ respectively. Hence $|\E\{F_m^{(k)}(\edge)|P_m\}|$ obeys the following recursive relation
\begin{equation}
	\begin{aligned}
		\left|\E\{F_m^{(k)}(\edge)|P_m\}\right|&\leq \left|\E\{F_m^{(k-1)}(\edge)|P_m\}\right|\\
		&+\gammadis^k\normw{\Delta_m (\cdot)}\\
		&+\gammadis^k\normw{\E\{F_m^{(0)}(\cdot)|P_m\}}.
	\end{aligned}
\end{equation}
By iterating the previous equation we achieve the following bound
\begin{equation}\label{equ.Fk_relation}
\begin{aligned}
	&\left|\E\{F_m^{(k)}(\edge)|P_m\} \right|\\
	&\leq \sum_{l=0}^{k}\gammadis^l\normw{\E\{F_m^{(0)}(\cdot)|P_m\}}+\sum_{l=1}^{k}\gammadis^l \normw{\Delta_m(\cdot)}\\
	&\leq \frac{1}{1-\gammadis}\left(\normw{\E\{F_m^{(0)}(\cdot)|P_m\}}+\gammadis \normw{\Delta_m(\cdot)}\right),
\end{aligned}
\end{equation}
where we have used the relation $\sum_{l=0}^{\infty}\gammadis^l=\frac{1}{1-\gammadis}$ to obtain a bound that  is independent of both $\edge$ and $k$.

Notice that $F_m^{(0)}(\edge)$ corresponds to the kind of update term encountered in the single-step algorithm of SARSA. It has been proven in \cite{singh2000convergence} as part of the convergence proof of the SARSA method that $F_m^{(0)}$ satisfies 
\begin{equation}\label{equ.F0_relation}
\normw{\E\{F_m^{(0)}(\cdot)|P_m\}}\leq \gammadis \normw{\Delta _m (\cdot)}+d_m,
\end{equation}
where $d_m$ converges to zero w.p.1. as $m\rightarrow \infty$. Here we recall from \cite{singh2000convergence} the main mathematical steps to prove Eq.~(\ref{equ.F0_relation}) as they will be useful later, when we consider the general scenario with time-dependent policies. The expected value of $F_m^{(0)}(\edge)$ can be written explicitly as
\begin{equation}\label{equ.F0_relation_2}
	\begin{aligned}
		&\E\{F_m^{(0)}(s,a)|P_m\}\\
		&\;=r(s,a)+\gammadis \sum_{s'}\Pr(s'|s,a)\sum_{a'}\Pr\left(a'|s',P_m\right)\htilde_m(s',a')\\
		&\phantom{=}\,-\qstar(s,a)\\
		&\;=f_m(s,a)+\gammadis \sum_{s'}\Pr(s'|s,a)g_m(s'),\\
	\end{aligned}
\end{equation}
where we have defined 
\begin{equation}
\begin{aligned}
f_m(s,a)&=r(s,a)+ \gammadis \sum_{s'} \Pr\left(s'|s,a\right)\max_b \{\tilde{h}_m (s',b)\} \\
&\quad-\qstar(s,a),\\
g_m(s')&=\sum_{a'}\Pr\left(a'|s',P_m\right)\htilde_m(s',a')-\max_b \{ \htilde_m (s',b) \}.
\end{aligned}
\end{equation}
The first term given above corresponds to the update term encountered in Q-learning algorithms and it has been proven to be bounded by
\begin{equation}\label{equ.f_bound}
|f_m(s,a)|\leq \gammadis \normw {\Delta_m (\cdot)},\; \forall s,a.
\end{equation}

In order to bound $g_m(s)$, let us recall that the policy (under the assumption that it is kept constant during an episode) is given by 
\begin{equation}\label{equ.policyB21}
\Pr(a|s,P_m)=\frac{\exp[\beta_m \htilde_m(s,a)]}{\sum_b\exp[\beta_m \htilde_m(s,b)]}.
\end{equation}
By simple algebraic manipulation, one can see that
\begin{equation}\label{equ.g_bound}
	g_m(s)<\frac{n_a}{\beta_m},\;\forall s,
\end{equation}
where $n_a=||\mathcal{A}||$ is the number of actions (assumed finite). Due to the fact that $\beta_m\rightarrow \infty$ as $ m\rightarrow \infty$ it follows that $g_m(s)$ converges to zero w.p.1. Eq.~(\ref{equ.F0_relation}) is recovered by plugging Eqs.~(\ref{equ.g_bound}) and (\ref{equ.f_bound}) into Eq.~(\ref{equ.F0_relation_2}) and defining $d_m=\gammadis n_a/\beta_m$. Finally, replacing Eq.~(\ref{equ.F0_relation}) into Eq.~(\ref{equ.Fk_relation}) we obtain the desired bound
\begin{equation}\label{equ.desired_bound}
\normw{\E\{F_m^{(k)}(\cdot)|P_m\}}\leq \frac{2\gammadis}{1-\gammadis}\normw{\Delta_m(\cdot)}+c_m\quad\forall k,
\end{equation}
where $c_m\equiv \frac{1}{1-\gammadis}d_m$ also converges to zero w.p.1. 

Eq.~(\ref{equ.desired_bound}) proves the contraction property for the case where the policy is not updated during episodes. Below we discuss the general case, where the policy may change every time step. As we will see the only difference with respect to the case discussed above is that an additional term must be added to the right hand side of Eq.~(\ref{equ.desired_bound}). Since this additional term converges to zero w.p.1 the contraction property also holds in that case.

\subsubsection{Policy update at every time step}

When on line updates are considered the policy might change every time step. Therefore, the probabilities in Eq.~(\ref{equ.F_truncated_expected_2}) no longer depend exclusively on $P_m$ but rather also on the rewards received by the agent during the episode. 
However, since most of this probabilities are taken as common factors and upper bounded by one, one can verify that most of the previous derivations still hold in the case where policies are changed every time step. In fact the only point where the previous derivations have to be generalized is in Eq.~(\ref{equ.F0_relation_2}). A time-dependent generalization of $g_m(s)$ can be defined by
\begin{equation}
	g'^{(t)}_m(s)=\sum_{a} \pi_m^{(t)}(a|s)\htilde_m (s,a)-\max_{b} \{\htilde_m(s,b)\},
\end{equation}
where $t$ could be any time step in the interval $\mathcal{I}_m=[T_m+1, T_{m+1}]$, i.e. the interval of time steps between the beginning and the end of episode $m$, and the policy is now given by
\begin{equation}\label{equ.policy_app_2}
	\pi^{(t)}_m(a\mid s) =\frac{\exp[\beta_{m} \htilde^{(t)}(s,a)]}{\sum_{b\in\mathcal{A}}\exp[\beta_{m} \htilde^{(t)}(s,b)]}.
\end{equation}
It is easy to verify that, similarly as in Eq.~(\ref{equ.g_bound}), if $\forall t\in \mathcal{I}_m$,  $g'^{(t)}_m(s)$ is upper bounded by a sequence converging to zero, Eq.~(\ref{equ.F0_relation}) also holds for time-dependent policies and hence Eq.~(\ref{equ.desired_bound}) too. 

Note that the only difference between Eqs.~(\ref{equ.policyB21}) and (\ref{equ.policy_app_2}) is that in the former, the policy depends on the $\htilde _m$ values while in the latter on the $\htilde^{(t)}$, with $t\in\mathcal{I}_m$. The difference between these two, however, tends to zero w.p.1 as the number of episodes increase. Given that rewards are bounded in the sense that $R^{(t)}\leq \mathcal{B}_R$, $\forall t$ for certain constant $\mathcal{B}_R$, we have that
\begin{equation}\label{equ.h_value_difference}
	|\htilde_t(\edge)-\htilde_m (\edge)|\leq \frac{1}{N_m+1}\left[|\htilde_m(\edge)|+ \frac{\mathcal{B}_R}{1-\gammadis}\right]\leq \frac{C}{N_m+1},
\end{equation}
where $C$ is a constant. Since $N_m\rightarrow\infty$ w.p.1, the previous difference converges to zero. 

To exploit Eq.~(\ref{equ.h_value_difference}), we bound $|g_m^{(t)}(s)|$ in the following way
\begin{equation}
\begin{aligned}
	|g_m^{(t)}(s)|\leq &|\max_a\{\htilde_m(s,a)\}-\max_a\{\htilde^{(t)}(s,a)\}|\\
	&+|\max_a\{\htilde^{(t)}(s,a)\}-\sum_a \pi^{(t)}(a|s)\htilde^{(t)}(s,a)|\\
	&+\sum_a\pi^{(t)}(a|s)|\htilde^{(t)}(s,a)-\htilde_m(s,a)|.
\end{aligned}
\end{equation}
It follows from Eq.~(\ref{equ.h_value_difference}) that the first and third line in the equation above are each upper bounded by $C/(N_m+1)$. Furthermore, the second term can be upper bounded by $n_a/\beta_m$ in the exact same way as in Eq.~(\ref{equ.g_bound}). Thus, 
\begin{equation}
	|g'^{(t)}_m|\leq \frac{n_a}{\beta_m}+2\frac{C}{N_m+1},
\end{equation}
which converges to zero w.p.1. This implies that Eq.~(\ref{equ.F0_relation_2}) also holds in the general case where the policy is updated every time step. This completes the proof of Eq.~(\ref{equ.desired_bound}).

\section{\label{sect:app_3} GLIE policy}
In this appendix, we consider a policy given by the probabilities 
\begin{equation}\label{equ.policy_app}
	\mathrm{Pr}^{(t)}_m(a\mid s) =\frac{\exp[\beta_{m} \htilde^{(t)}(s,a)]}{\sum_{b\in\mathcal{A}}\exp[\beta_{m} \htilde^{(t)}(s,b)]},
\end{equation}
and derive conditions on the coefficients $\beta_m$ in order to guarantee that this policy is GLIE on the $\htilde$-values. We closely follow the derivation provided in \cite{singh2000convergence}, and therefore will omit most of the details. Here, however, we focus on episodic tasks, in contrast to the continuous time tasks considered in \cite{singh2000convergence}. Moreover, the coefficients $\beta_m$ will depend exclusively on the episode index $m$, instead of the state $s$. In this way we preserve a local model in the sense defined in the main text, where the $\beta$ coefficient can be updated using exclusively environmental signals (in this case, the end of an episode).

Letting $\beta_m\rightarrow \infty$ is enough to guarantee that the policy is greedy in the limit. However, the speed at which $\beta_m$ grows as a function of $m$ must be upper bounded in order to additionally guarantee infinite exploration. In the following we derive such a bound. 

Let us denote as $\mathrm{Pr}_m(s,a)$ the probability that during episode $m$ the state-action pair $(s,a)$ is visited at least once. Hence, infinite exploration occurs if $\forall s,a$
\begin{equation}\label{equ.infinite_probabilities}
	\sum_{m=1}^{\infty}\mathrm{Pr}_m(s,a)=\infty.
\end{equation}
Considering that $\sum _{m=1}^{\infty} c/m=\infty$ for any constant $c$, as a consequence of the Borel Cantelli lemma we have that a sufficient condition for Eq.~(\ref{equ.infinite_probabilities}) is given by 
\begin{equation}\label{equ.borel_bound}
	\mathrm{Pr}_m(s,a)\geq c/m,
\end{equation}
for some constant $c$. Therefore we would like to pick a bound on $\beta_m$ in such a way that Eq.~(\ref{equ.borel_bound}) is satisfied. 

Let us denote by $\mathrm{Pr}_m (s)$ the probability that during episode $m$ state $s$ is visited at least once and let  
\begin{equation}
	p_{\mathrm{min}}(m)=\min_{a,s,t\in \mathcal{I}_m}\left\{\mathrm{Pr}^{(t)}_m(a|s)\right\}
\end{equation}
be the minimum probability to take any action at any time step during the $m$th episode. It follows that
\begin{equation}\label{equ.probability_state_action}
	\mathrm{Pr}_m(s,a)\geq \mathrm{Pr}_m(s) p_{\mathrm{min}}(m).
\end{equation}
The first factor can in turn be bounded by a function of $p_{\mathrm{min}}(m)$ by noting the following. In a communicating MDP, any state can be reached from any other non-terminal state with nonzero probability. That means that independently of the initial state of the $m$th episode, there exists a sequence of actions that lead to any state $s$ with some nonzero probability $p_0$. Such probability is constant and it is given by a product of transition probabilities of the MDP. In the worst case scenario, it could happen that $s$ can only be reached by taking a specific sequence of actions that leads the agent to visit all non terminal states before reaching $s$. Hence $\mathrm{Pr}_m(s)$ can be bounded by the product of probabilities to pick those actions weighted by the transition probabilities of such a sequence. Given that the probability to take any action is in turn lower bounded by $p_{\mathrm{min}}(m)$, we conclude that  
\begin{equation}\label{equ.probability_state}
	\mathrm{Pr}_m(s)\geq p_0\, \left[ p_{\mathrm{min}}(m)\right]^{n_s-1},
\end{equation}
where $n_s=||\mathcal{S}||-||\mathcal{S}_T||$ is the number of non terminal states.

In order to bound $p_{\mathrm{min}}(m)$ note that 
\begin{equation}
 \begin{aligned}
	 \mathrm{Pr}_m^{(t)}(a|s)&\geq \frac{1}{n_a} \exp\left[-\max_{a,b}\left\{\htilde^{(t)}(a|s)-\htilde^{(t)}(b|s)\right\}\beta_m \right]\\
	 &\geq \frac{1}{n_a}\exp\left[-2\mathcal{B}_{\tilde{h}}\beta_m \right],
 \end{aligned}
\end{equation}
where $\mathcal{B}_{\htilde}\geq \htilde^{(t)}(s,a), \forall s,a$ is an upper bound that exists because the rewards are bounded.
Hence, it follows that $p_{\mathrm{min}}(m)\geq \exp\left[-2 \mathcal{B}_{\htilde} \beta_m\right]/n_a$. Replacing this inequality in Eq.~(\ref{equ.probability_state}) and using Eq.~(\ref{equ.probability_state_action}) we get that
\begin{equation}
	\begin{aligned}
		\mathrm{Pr}_m(s,a)&\geq p_0 \,p_{\mathrm{min}}^{n_s}(m)\\
		&\geq \frac{p_0}{n_a^{n_s}}\exp\left[-2n_s\mathcal{B}_{\htilde} \beta_m \right].
	\end{aligned}
\end{equation}
Therefore, by choosing $\beta_m$ in such a way that 
\begin{equation}
	\beta_m \leq \frac{1}{2n_s\mathcal{B}_{\htilde}}\ln(m),
\end{equation}
Eq.~(\ref{equ.borel_bound}) holds, and thus the policy is guaranteed to preserve infinite exploration of all state-action pairs.

\phantomsection \label{references} 

\addcontentsline{toc}{section}{\protect\numberline{\Alph{section}}References} 
\bibliographystyle{ieeetr}
\textbf{References}
\bibliography{manuscript}

\end{document}